\documentclass{article}

\usepackage{amsmath}
\usepackage{amsfonts}
\usepackage{booktabs} 
\usepackage{color} 
\usepackage{microtype}
\usepackage{graphicx}
\usepackage{subfigure}
\usepackage{booktabs} 
\usepackage{tabularx}
\usepackage{algorithm}
\usepackage{algorithmic}
\usepackage{url}
\usepackage{relsize}
\usepackage{hyperref}
\usepackage{subfigure}
\usepackage[english]{babel}
\usepackage{blindtext}
\usepackage[group-separator={,}]{siunitx}
\usepackage{multicol}
\usepackage{array}
\usepackage{frame}
\usepackage{adjustbox}
\usepackage[percent]{overpic}
\newcolumntype{C}{>{\lower.2ex\hbox to 2ex\bgroup\hss}c<{\hss\egroup}}

\definecolor{silver}{cmyk}{0,0,0,0.3}
\definecolor{yellow}{cmyk}{0,0,0.9,0.0}
\definecolor{reddishyellow}{cmyk}{0,0.22,1.0,0.0}
\definecolor{black}{cmyk}{0,0,0.0,1.0}
\definecolor{darkYellow}{cmyk}{0.2,0.4,1.0,0}
\definecolor{darkSilver}{cmyk}{0,0,0,0.1}
\definecolor{grey}{cmyk}{0,0,0,0.5}
\definecolor{darkgreen}{cmyk}{0.6,0,0.8,0}

\newcommand{\w}{\omega}
\newcommand{\triplet}{(i,j,k)}

\usepackage{hyperref}

\usepackage[accepted]{icml2018}
\usepackage[font=footnotesize,skip=2pt]{caption}

\icmltitlerunning{Dimensionality reduction using triplets}

\begin{document}

\twocolumn[
\icmltitle{A more globally accurate\\
dimensionality reduction method using triplets}




\begin{icmlauthorlist}
\icmlauthor{Ehsan Amid}{to}
\icmlauthor{Manfred K. Warmuth}{to}
\end{icmlauthorlist}

\icmlaffiliation{to}{Department of Computer Science, University of California, Santa Cruz}

\icmlcorrespondingauthor{Ehsan Amid}{eamid@ucsc.edu}
\icmlcorrespondingauthor{Manfred K. Warmuth}{manfred@ucsc.edu}


\vskip 0.3in
]



\printAffiliationsAndNotice{}

\begin{abstract}
We first show that the commonly used dimensionality reduction
(DR) methods such as t-SNE and LargeVis
poorly capture the global structure of the data
in the low dimensional embedding.
We show this via a number of tests for the DR
methods that can be easily applied by any practitioner to the dataset at hand. 
Surprisingly enough, t-SNE performs the best 
w.r.t. the commonly used measures that reward
the local neighborhood accuracy such as precision-recall 
while having the worst performance in our tests for global structure.
We then contrast the performance of these two DR method
against our new method called {\em TriMap}. 
The main idea behind TriMap is to capture higher orders of structure 
with triplet information (instead of pairwise information
used by t-SNE and LargeVis), and to minimize a robust loss
function for satisfying the chosen triplets.
We provide compelling experimental evidence on large natural datasets 
for the clear advantage of the TriMap DR results. As
LargeVis, TriMap scales linearly with the number of data points.
\end{abstract}

\section{Introduction}

Information visualization using dimensionality reduction
(DR) is a fundamental step for gaining insight about a dataset.
Motivated by the fact that the humans essentially think in two or three dimensions, DR has been studied extensively throughout the years~\cite{isomap,som, lle, sne, tsne}. However, choosing the best DR method to visualize a certain dataset is yet an unspecified task. Despite the plethora of quantitative measures such as trustworthiness-continuity~\cite{tc}, mean (smoothed) precision-recall~\cite{nerv}, and nearest-neighbor accuracy~\cite{review}, there is no standard procedure to assess the quality of a low-dimensional embedding in general. Also these measures only focus on the local neighborhood of
each point in the original space and low-dimensional
embedding and fail to reflect global aspects of the
data. We will argue that selecting a DR method based on
these local measures is misleading.

In this paper, we mainly focus on the global properties of the data. 
These properties include the relative placements of the
clusters in the dataset as well as revelation of outlier
points that are located far away from the rest of the
points in the high-dimensional space.  It appears that
quantifying the global properties is significantly more
complicated and local quality measures of DR (such as precision-recall\footnote{We tried other measures such as Trustworthiness-Continuity on toy examples, but these measures can be extremely expensive to calculate for large datasets.})
do not capture these properties. Thus we resort to an evaluation based 
on visual clues that allows the practitioner to build
confidence about the obtained DR results. 

To this end, we propose a number of transformations that happen naturally to 
real-world datasets. We argue that any DR method that claims to
preserve the global structure of the data,
should be able to handle these transformations. 
\citet{ackerman} took a similar more theoretical approach for the task of clustering where certain natural properties are used to taxonomize different clustering algorithms.

Next, we introduce a new DR method, called TriMap. The main idea behind TriMap is to capture higher orders of structure in the data by 
considering the relative similarities of a chosen set of triplets of points. 
The method minimizes a loss over triplets that measures
how well a triplet is satisfied. This loss is made robust
by capping it with a damping function. 
We provide compelling experimental results on large natural datasets. 
We show that t-SNE and LargeVis, which are both based on preserving the pairwise (dis)similarities of the points, tend to form spurious clusters with clean boundaries and fail to properly reflect the global aspects of the data. 
On the other hand, TriMap's DR results preserve more of the 
global structure of the data as well as reveals outliers. 

\begin{figure*}[th!]
\begin{center}
\begin{tabular}{m{3mm} m{0.22\textwidth} m{0.22\textwidth} m{0.22\textwidth}}
\multicolumn{1}{c}{ }  & \multicolumn{1}{c}{t-SNE} & \multicolumn{1}{c}{LargeVis} & \multicolumn{1}{c}{TriMap}\\\hline
$\star$ &
\begin{center}\begin{overpic}[width=0.22\textwidth]{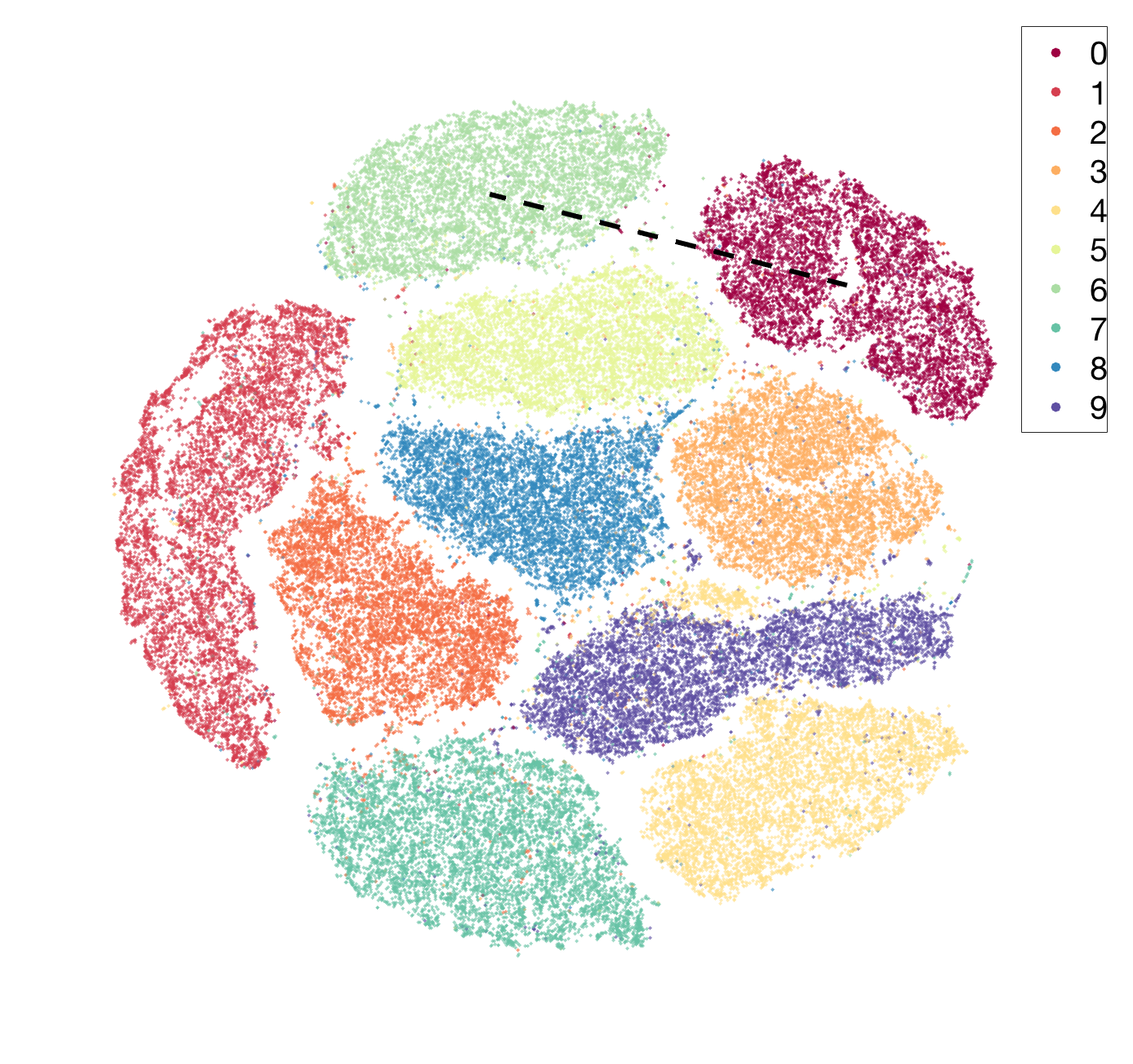}
 \put (32,-5) {\small AUC = $0.130$}
\end{overpic}\end{center}
& \begin{center}\begin{overpic}[width=0.22\textwidth]{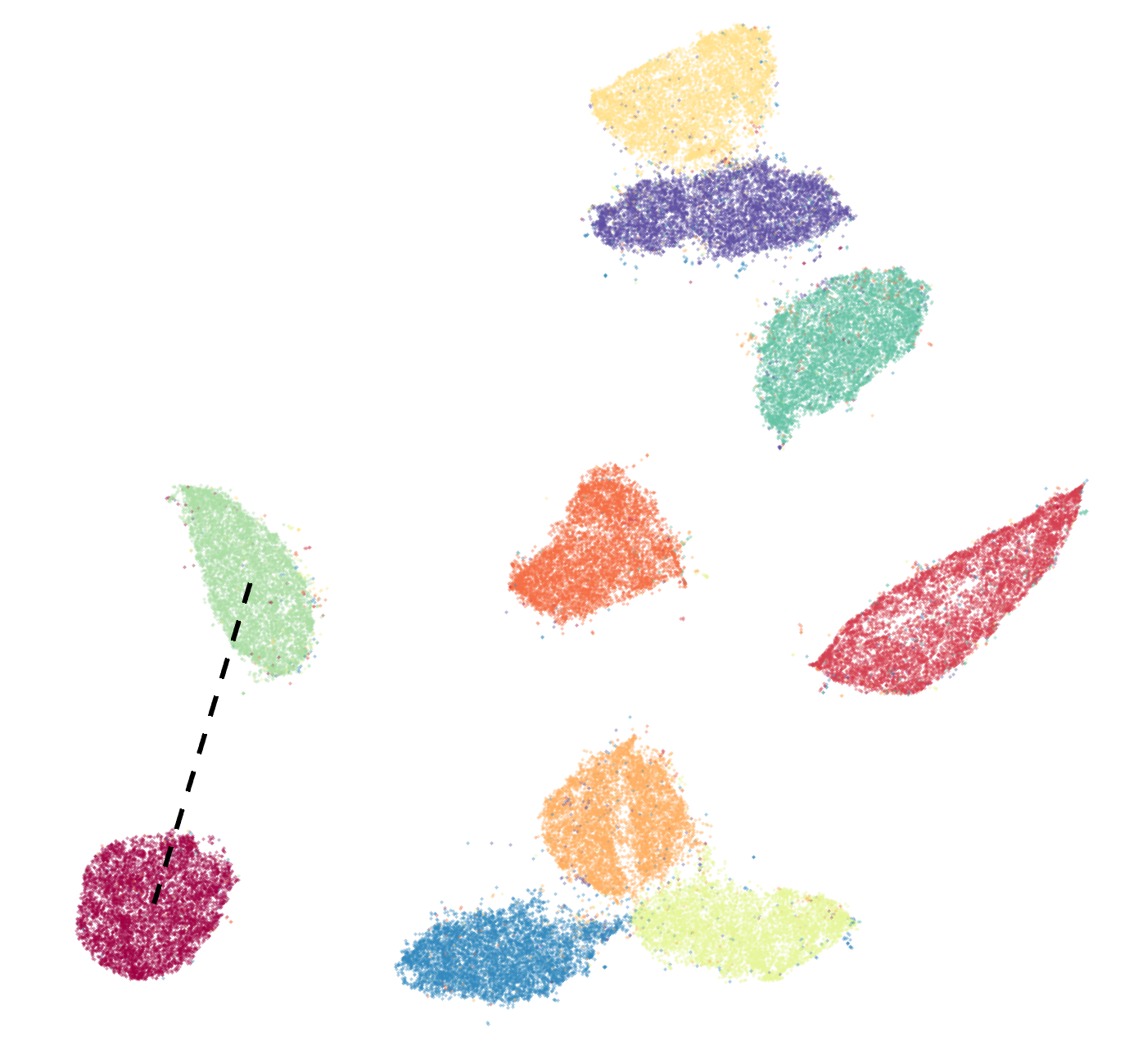}
 \put (32,-5) {\small AUC = $0.062$}
\end{overpic}\end{center}
& \begin{center}\begin{overpic}[width=0.22\textwidth]{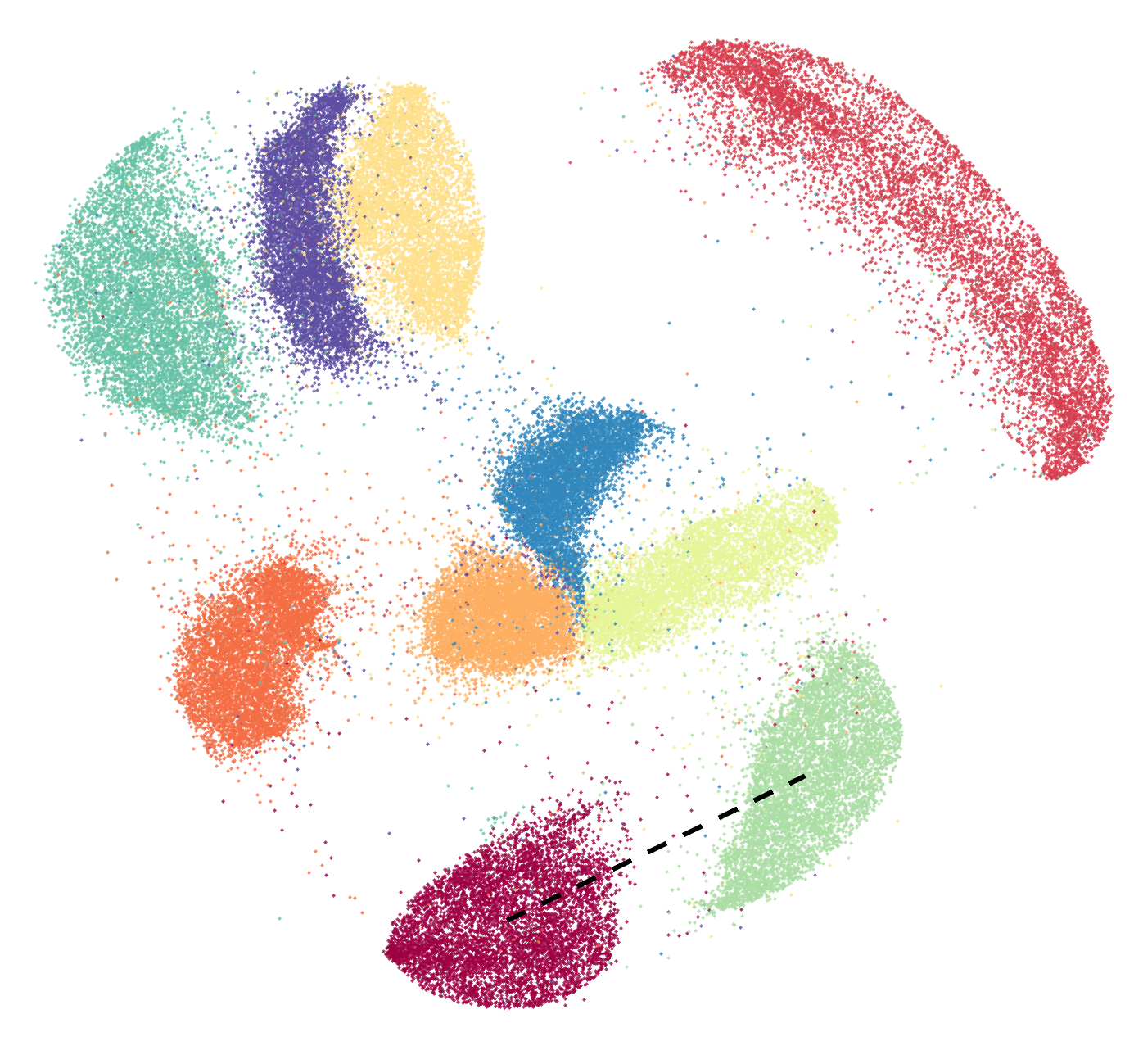}
 \put (32,-5) {\small AUC = $0.036$}
\end{overpic}\end{center}\\\hline\hline
{\small 2.1.a} & \begin{center}\begin{overpic}[width=0.22\textwidth]{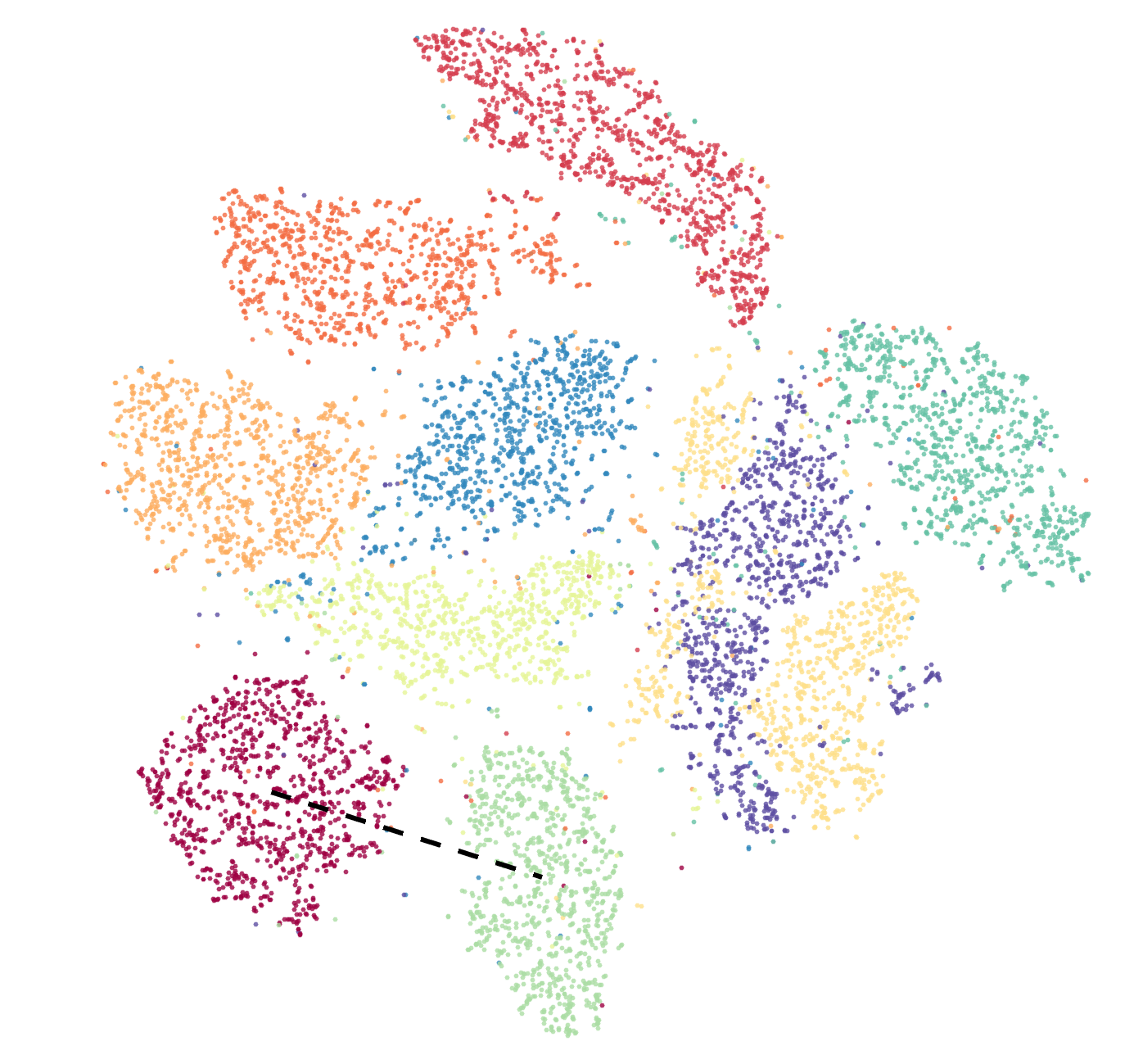}
 \put (32,-5) {\small AUC = $0.321$}
 \end{overpic}\end{center}
 &   \begin{center}\begin{overpic}[width=0.22\textwidth]{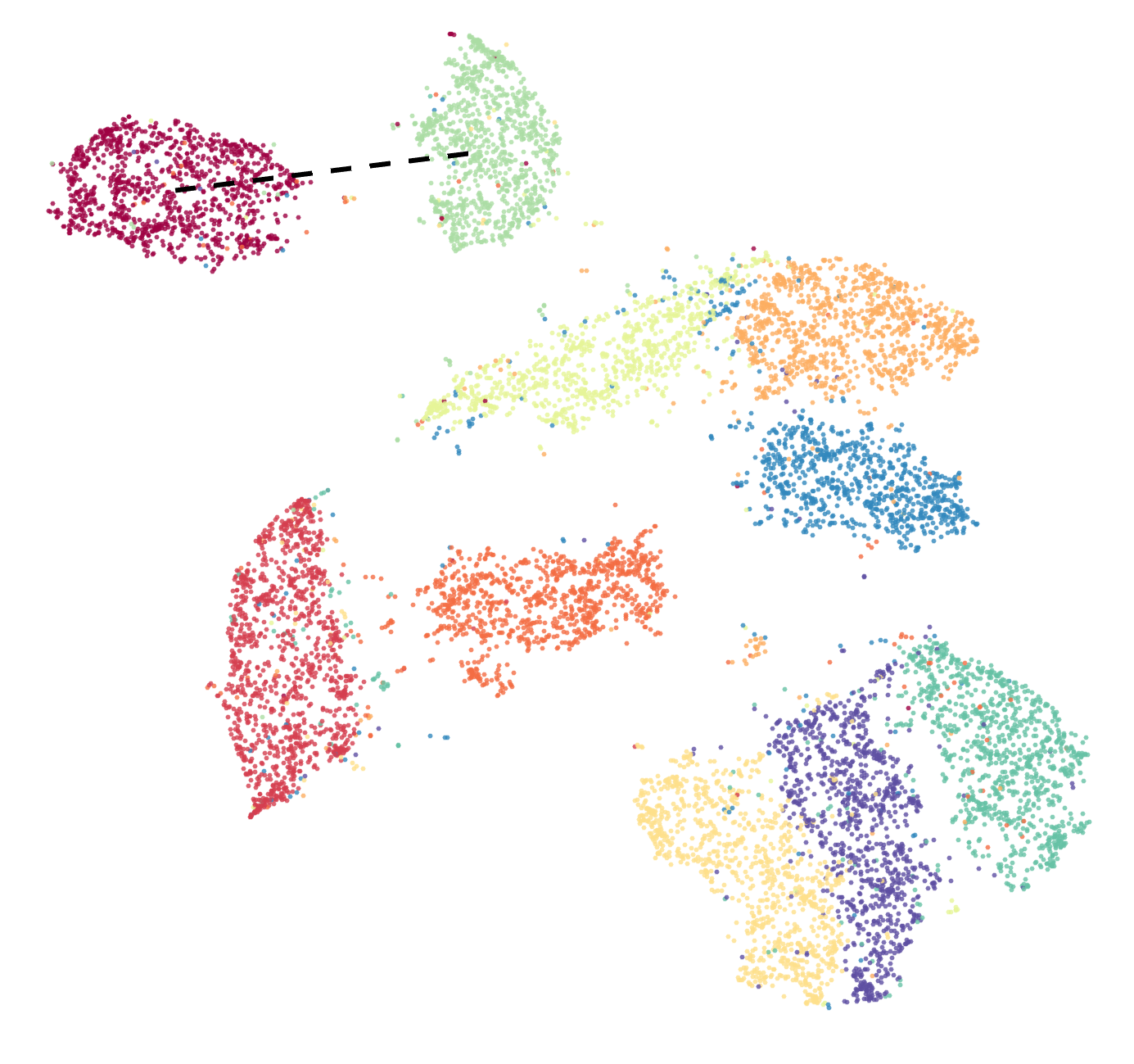}
 \put (32,-5) {\small AUC = $0.233$}
 \end{overpic}\end{center}
 & \begin{center}\begin{overpic}[width=0.22\textwidth]{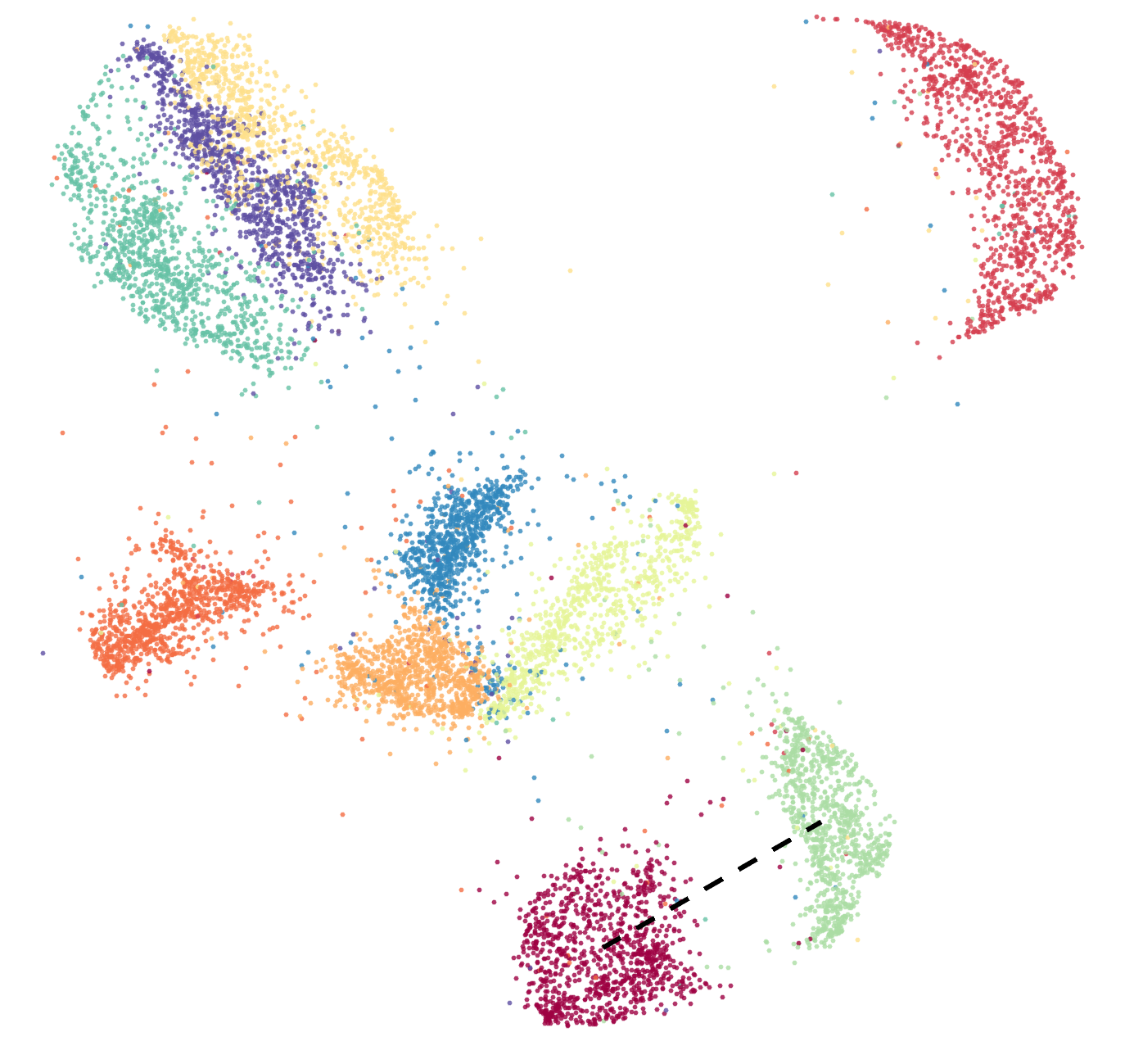}
 \put (32,-5) {\small AUC = $0.156$}
 \end{overpic}\end{center}\\\hline
{\small 2.1.b} & \begin{center}\begin{overpic}[width=0.22\textwidth]{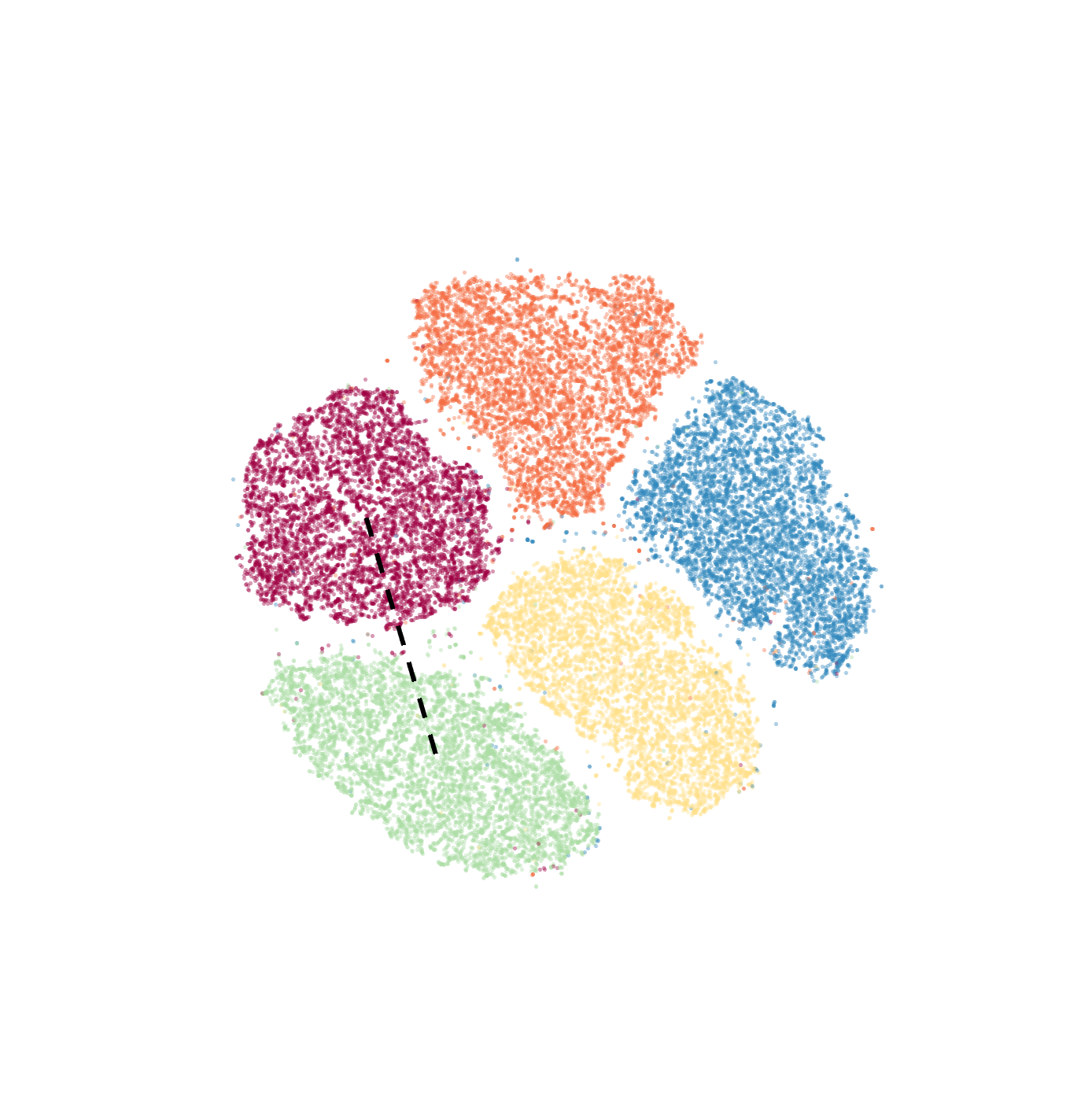}
 \put (32,-5) {\small AUC = $0.185$}
\end{overpic}\end{center}
& \begin{center}\begin{overpic}[width=0.22\textwidth]{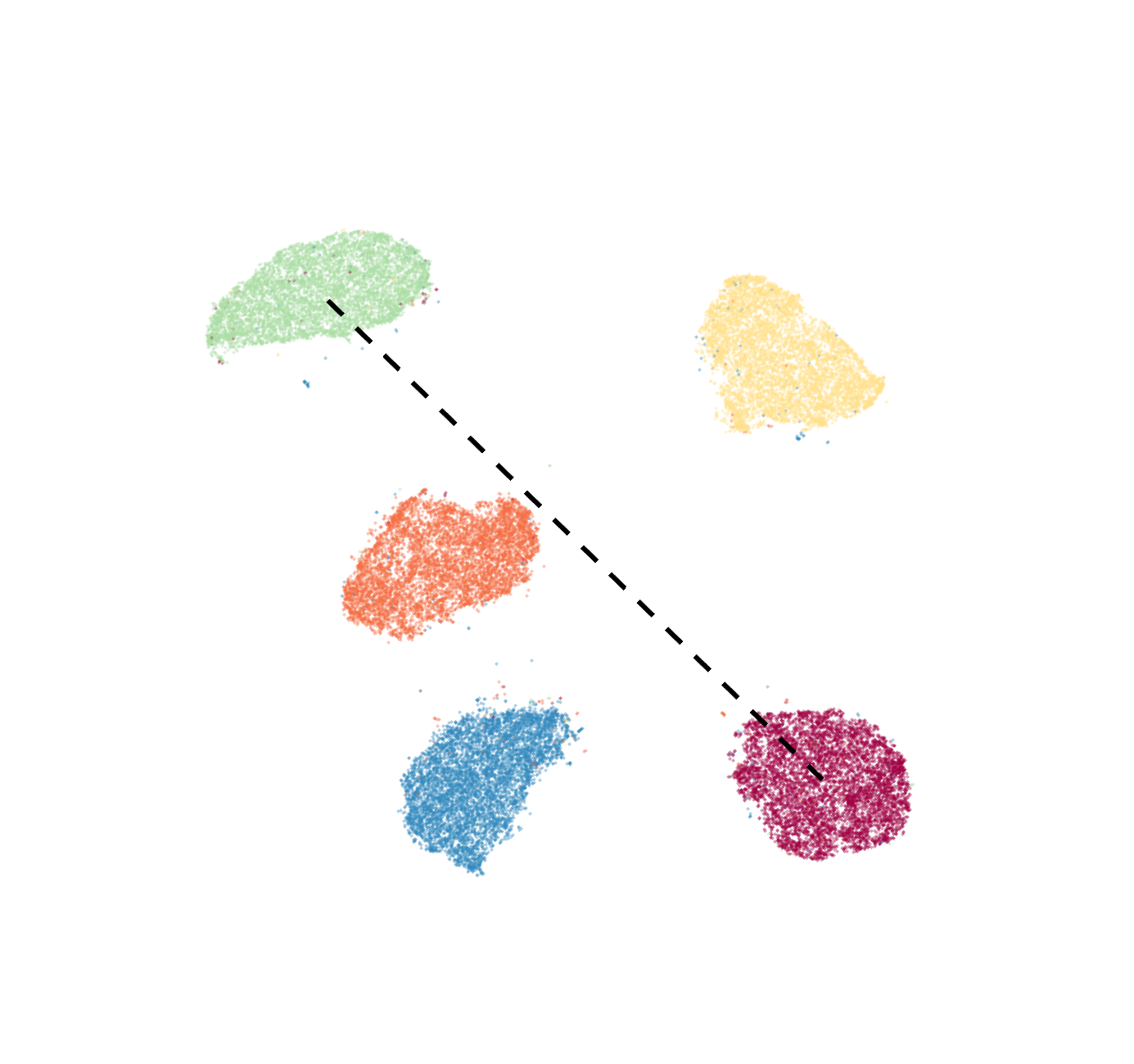}
 \put (32,-9) {\small AUC = $0.074$}
\end{overpic}\end{center}
& \begin{center}\begin{overpic}[width=0.22\textwidth]{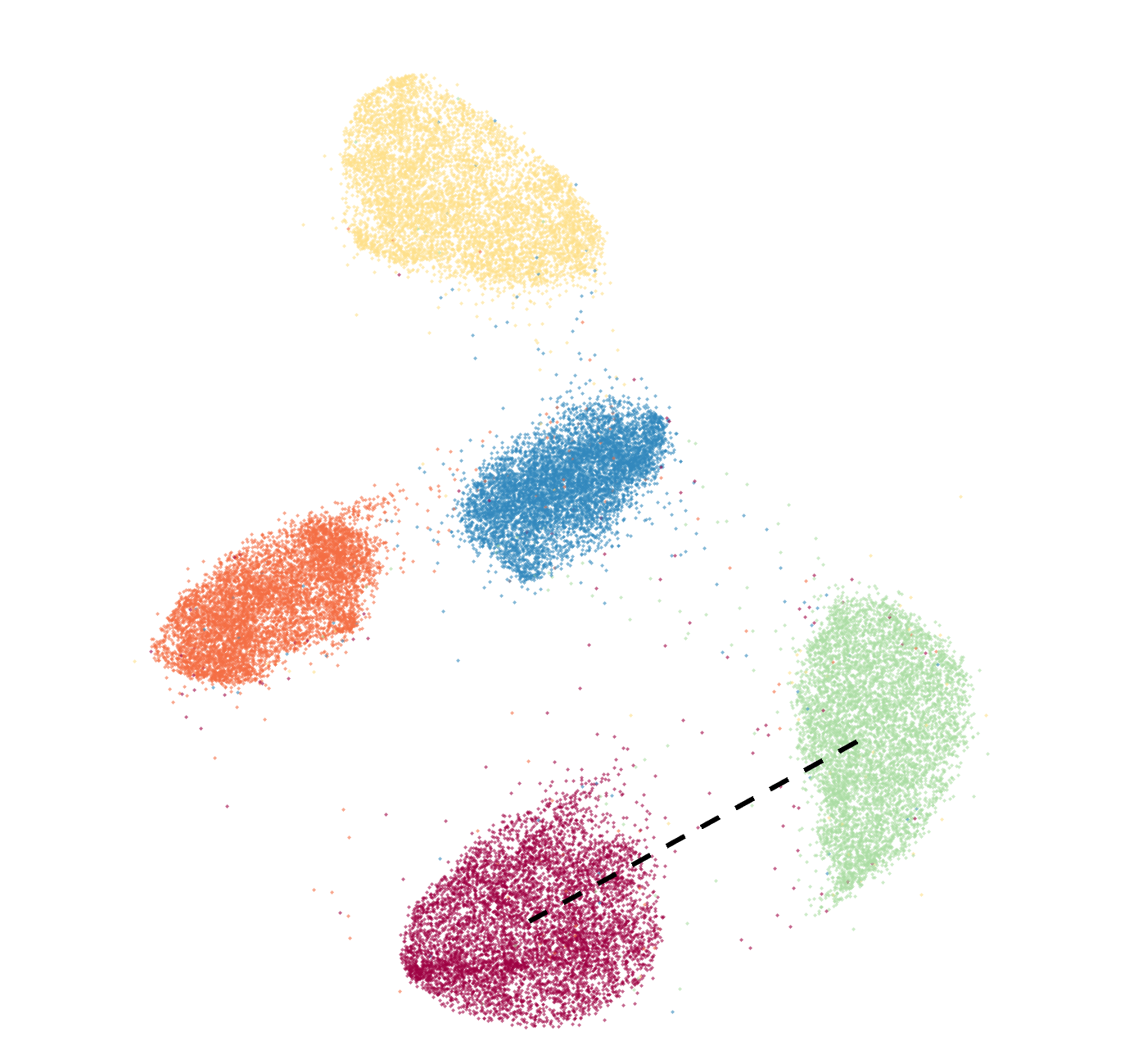}
 \put (32,-9) {\small AUC = $0.040$}
\end{overpic}\end{center}\\\hline
\end{tabular}
\caption{Dimensionality reduction tests: $\star$) full dataset, 2.1.a) a random $\% 10$ subset, 2.1.b) even digits only. The dotted line between the centers of the clusters `0' and `6' 
is drawn for comparing the pairwise distance between the
cluster centers before and after removing the subsets from
the dataset. Visualization best viewed in color.}\label{fig:tests}
\end{center}
\vspace{-0.7cm}
\end{figure*}

\renewcommand{\thefigure}{\arabic{figure} (Cont.)}
\addtocounter{figure}{-1}

\begin{figure*}[th!]
\begin{center}
\begin{tabular}{m{3mm} m{0.22\textwidth} m{0.22\textwidth} m{0.22\textwidth}}
\multicolumn{1}{c}{ }  & \multicolumn{1}{c}{t-SNE} & \multicolumn{1}{c}{LargeVis} & \multicolumn{1}{c}{TriMap}\\\hline
$\star$ & \begin{center}\begin{overpic}[width=0.22\textwidth]{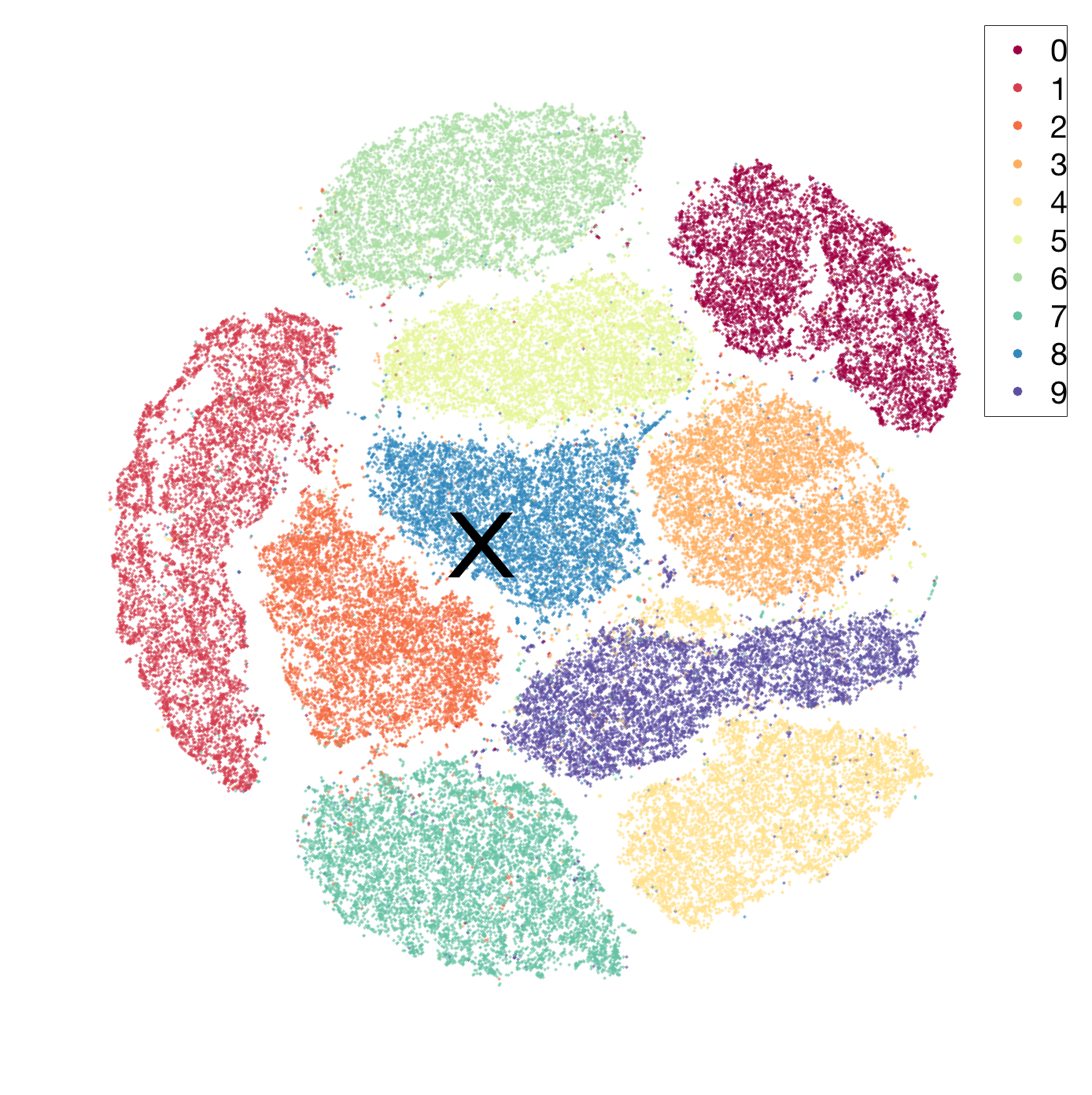}
 \put (32,-5) {\small AUC = $0.130$}
\end{overpic}\end{center}
&   \begin{center}\begin{overpic}[width=0.22\textwidth]{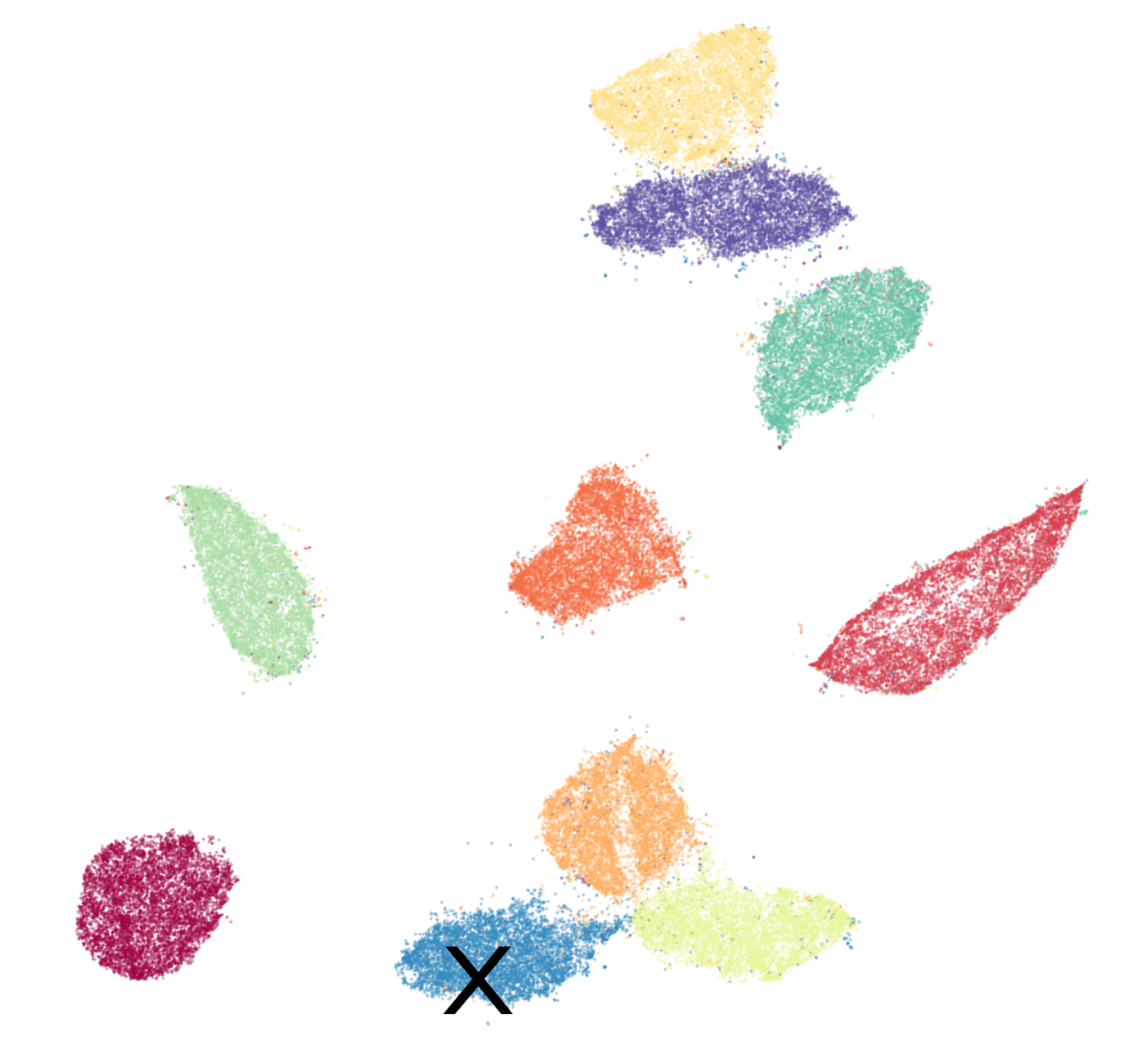}
 \put (32,-9) {\small AUC = $0.062$}
\end{overpic}\end{center}
& \begin{center}\begin{overpic}[width=0.22\textwidth]{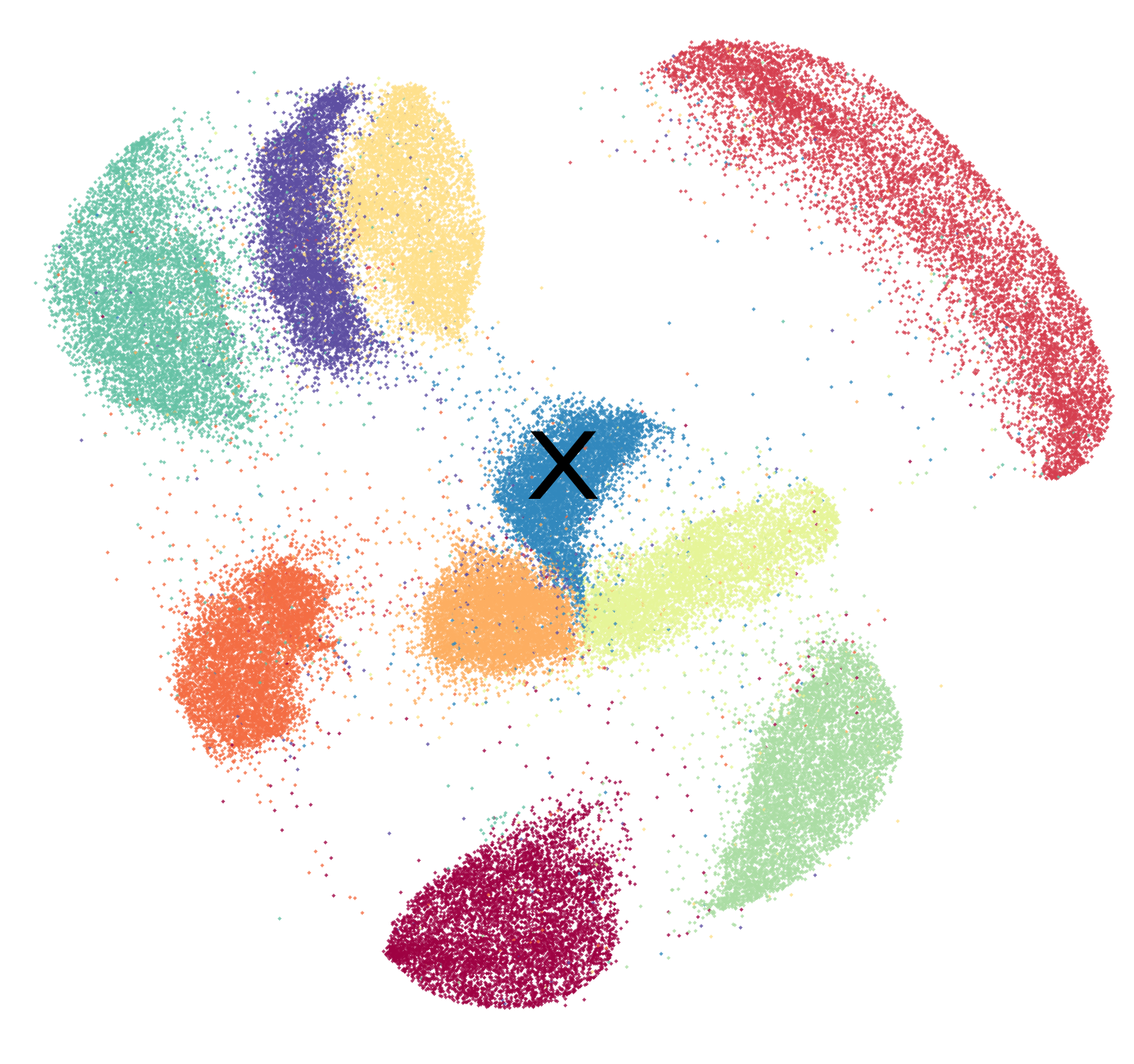}
 \put (32,-9) {\small AUC = $0.036$}
\end{overpic}\end{center}\\\hline\hline
{\small 2.2} & \begin{center}\begin{overpic}[width=0.22\textwidth]{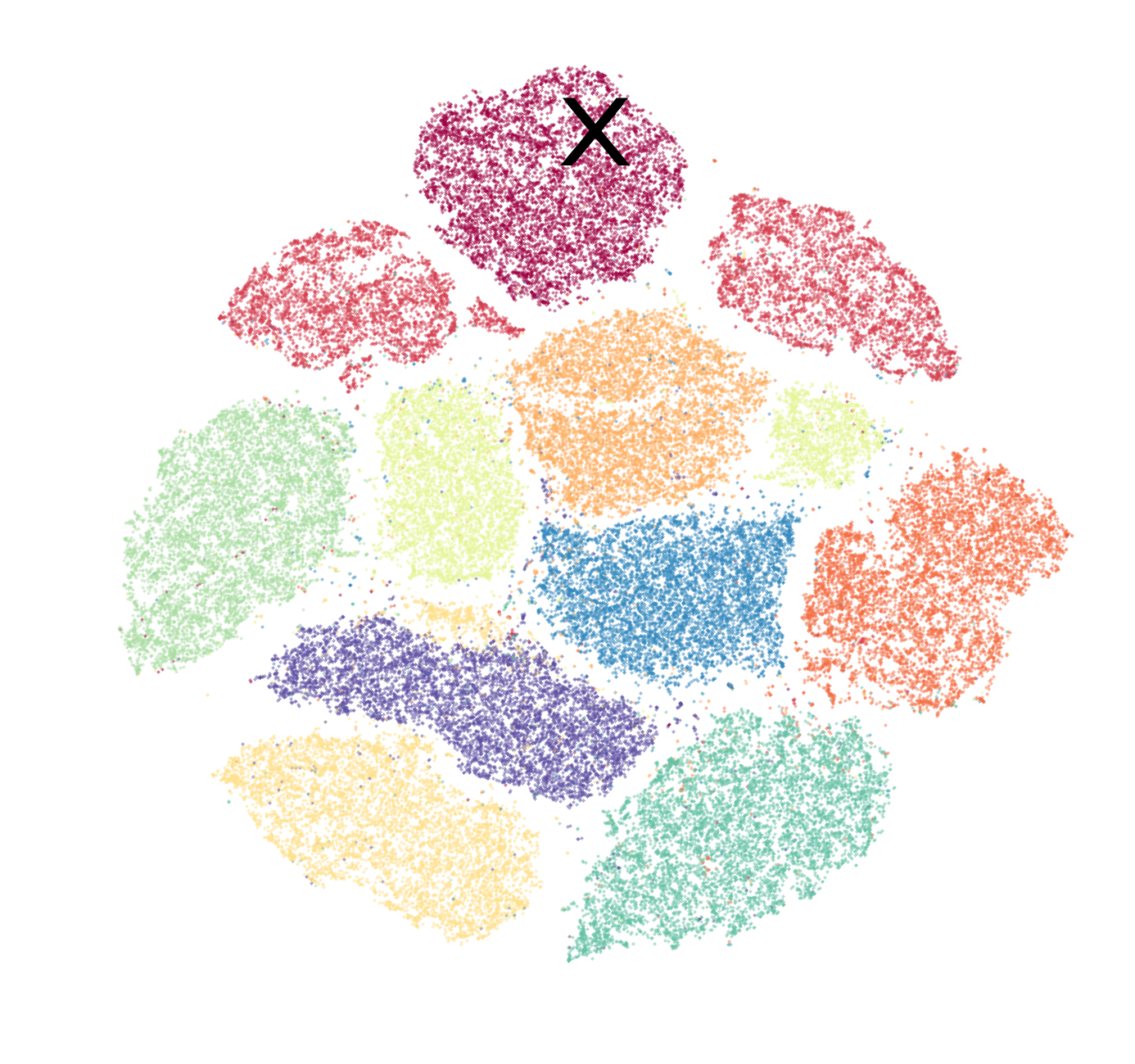}
 \put (32,-9) {\small AUC = $0.130$}
\end{overpic}\end{center}
& \begin{center}\begin{overpic}[width=0.22\textwidth]{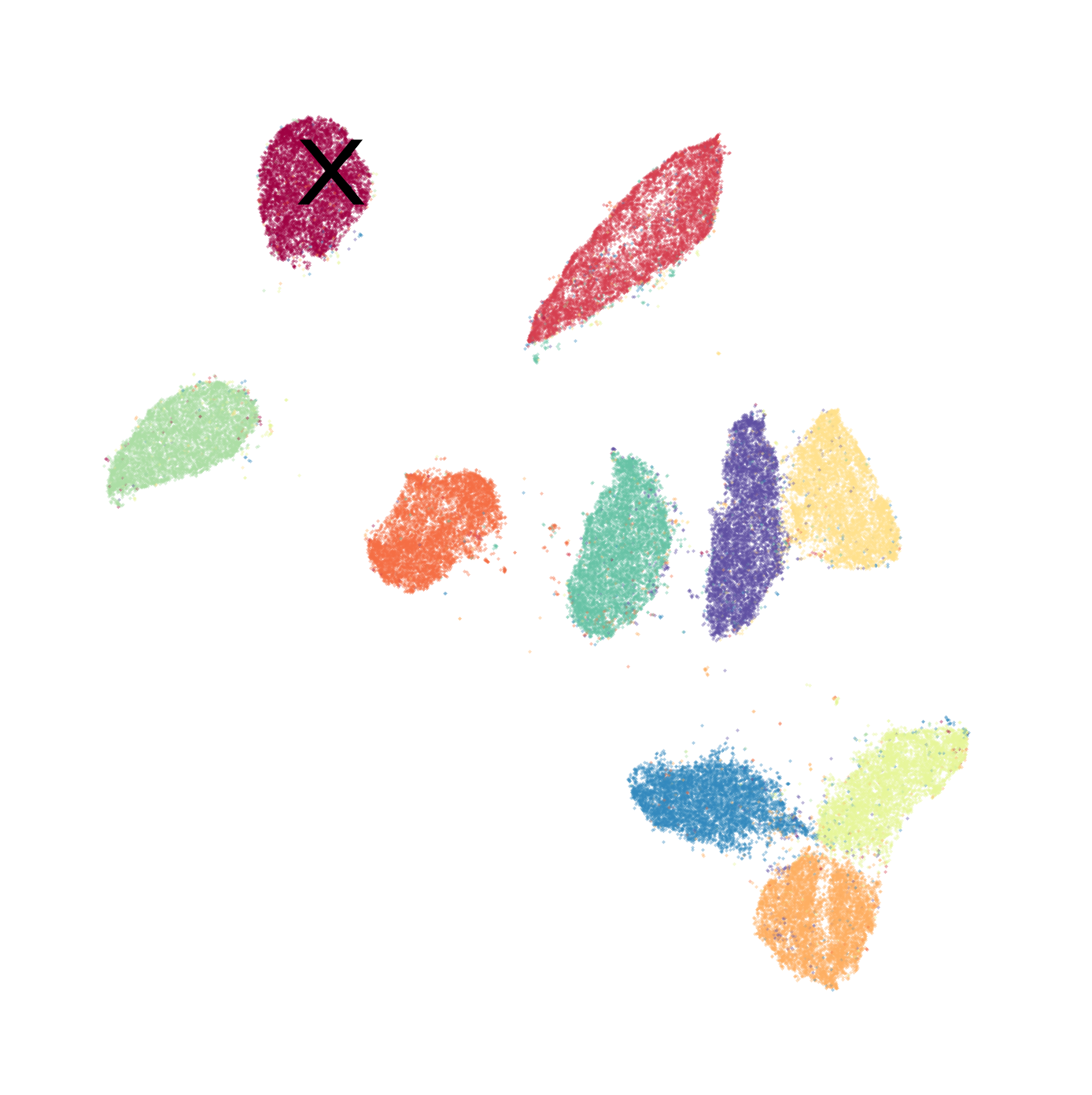}
 \put (32,-5) {\small AUC = $0.062$}
\end{overpic}\end{center}
& \begin{center}\begin{overpic}[width=0.22\textwidth]{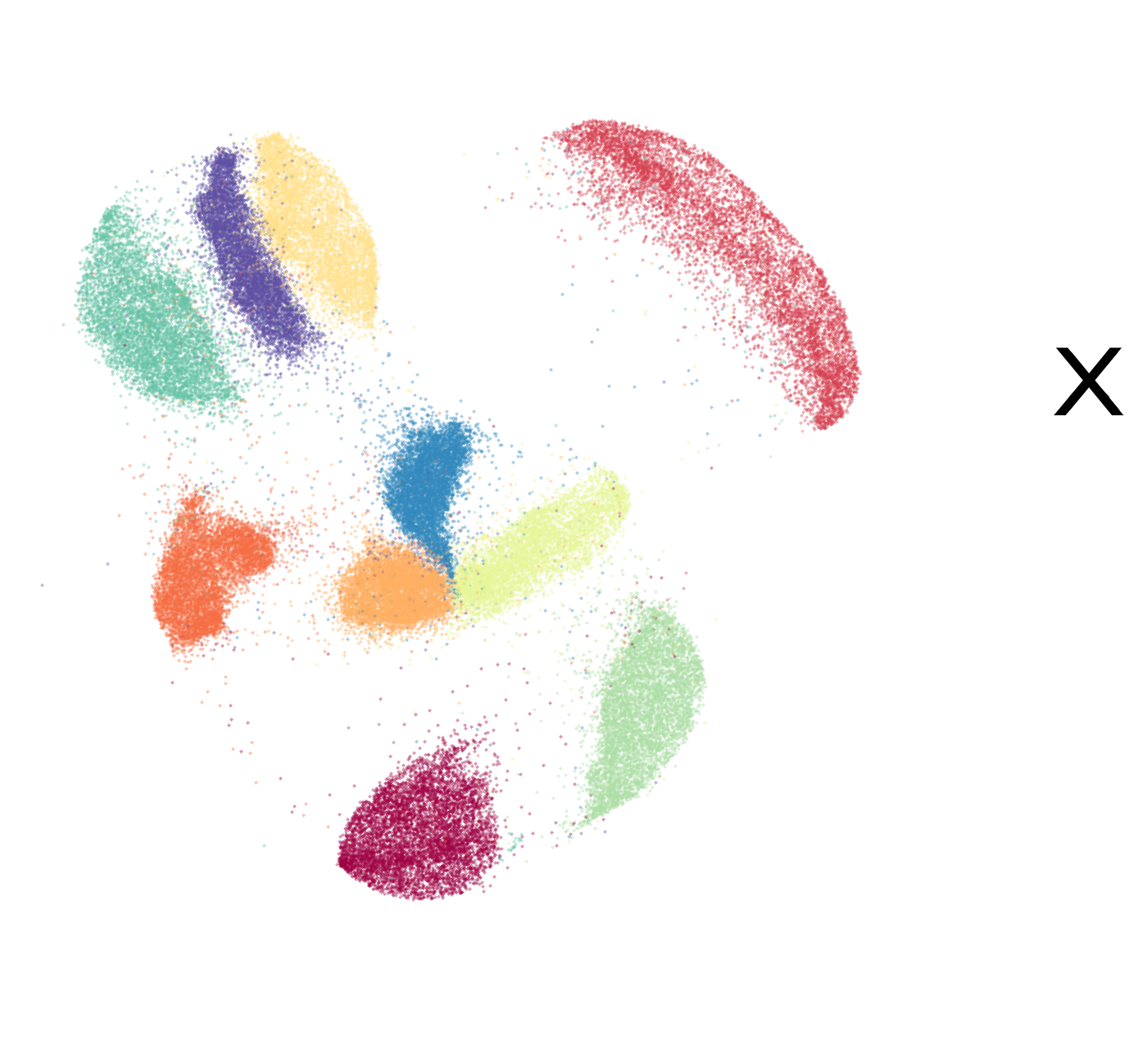}
 \put (32,-9) {\small AUC = $0.037$}
\end{overpic}\end{center}\\\hline
{\small 2.3} & \begin{center}\begin{overpic}[width=0.22\textwidth]{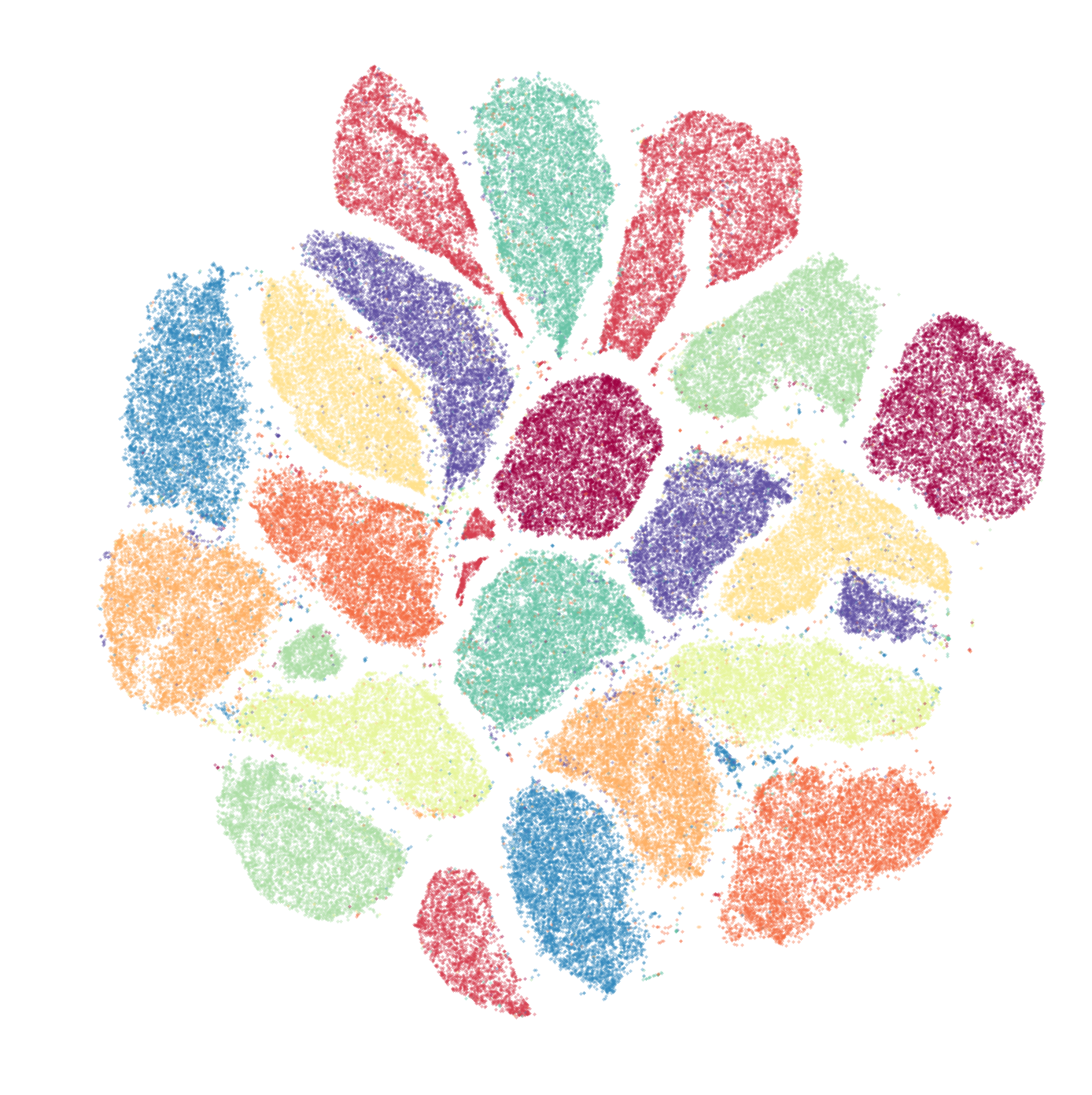}
 \put (32,-5) {\small AUC = $0.075$}
 \end{overpic}\end{center}
 & \begin{center}\begin{overpic}[width=0.22\textwidth]{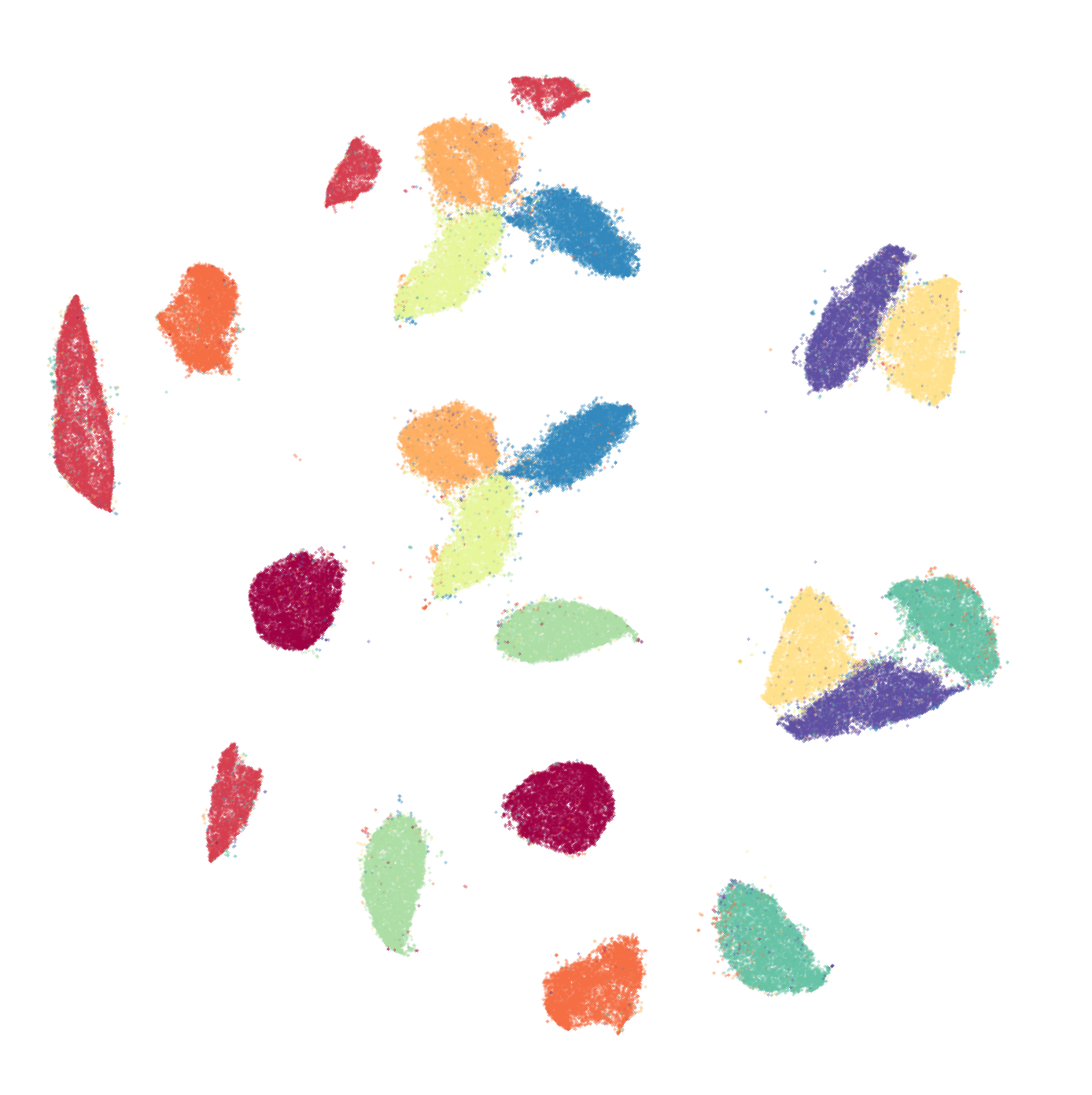}
 \put (32,-5) {\small AUC = $0.050$}
 \end{overpic}\end{center}
 & \begin{center}\begin{overpic}[width=0.22\textwidth]{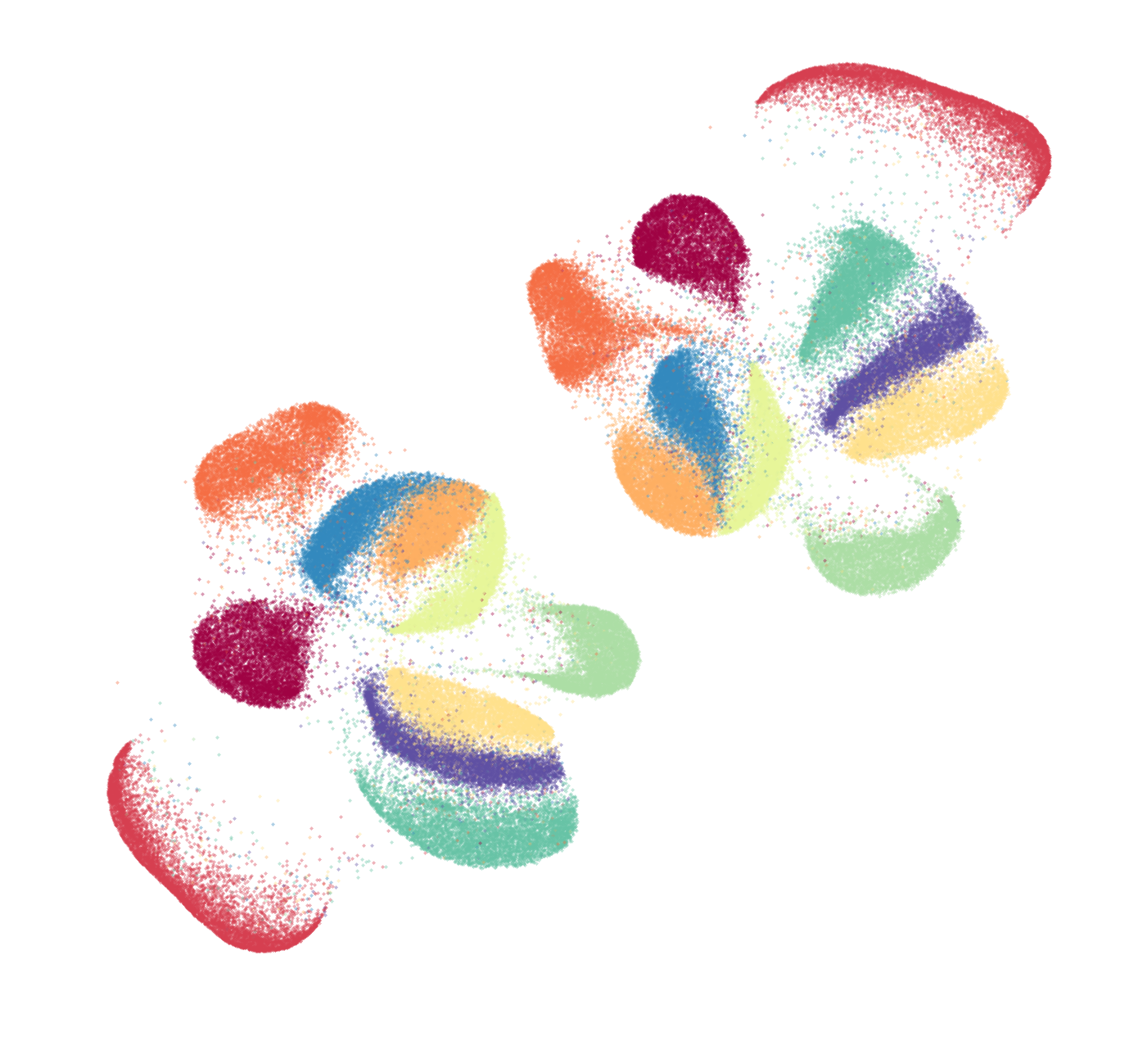}
 \put (32,-9) {\small AUC = $0.025$}
 \end{overpic}\end{center}\\\hline
\end{tabular}
\caption{Dimensionality reduction tests: $\star$) full dataset, 2.2) outlier, 2.3) multiple scales. Figures best viewed in color.}
\end{center}
\vspace{-0.7cm}
\end{figure*}

\renewcommand{\thefigure}{\arabic{figure}}

\section{Dimensionality reduction tests}

Our suggested tests for DR methods are given below. The tests mimic the natural scenarios that happen in the real-world datasets.
In each case we run the test on three DR methods:
t-SNE, which is currently the most widely used DR method;
LargeVis, a more recent method that is also based on pairwise
(dis)similarities; \& TriMap, our new method based on triplet information. 
We defer the description of t-SNE \& LargeVis 
as well as our new TriMap method to later sections.
In our experiments, we use fast t-SNE~\cite{bhtsne} for speed up.

As a running example, we perform the tests on the MNIST
dataset\footnote{\url{http://yann.lecun.com/exdb/mnist/}}
which contains \num{70000} images of handwritten digits
(represented by $784$ dimensions)\footnote{For ease of
comparison, we use the same initial solution for TriMap,
whenever possible. The current implementation of fast
t-SNE and LargeVis do not support initial
solutions.}. 
Figure~\ref{fig:tests}($\star$) shows the results on the full dataset
by the three methods.
The tests are repeated many times and the results are 
consistent over multiple runs for all methods. We normalize the data to
have pixel values between $[0,1]$. For each result, we
calculate the area under the curve (AUC) of the mean
precision-recall of the embedding \citep{nerv}.\footnote{%
More precisely, for each point we fix a neighborhood size of $20$
in the high-dimension as the ``relevant points'' and vary the
neighborhood size in the low-dimension between $1$ and
$100$. This gives us a precision and recall curve for each point. 
The mean curve is obtained by averaging the curves for all
points, and the AUC measure is the area under that curve.}

\subsection{Partial observation tests} 
A  DR tool should be invariant to removing a subset of
points from the dataset. 
The placement of the remaining points (or
clusters) after running the DR method on the reduced data
set should remain relatively unchanged compared to the embedding 
created from the full dataset.
In a first test, the subset of
removed points are selected at random.
This test mimics the fact that in reality we only see a 
sample from the data distribution and the DR method should
be stable w.r.t. the sample size.
Figure~\ref{fig:tests}(2.1.a) shows the results
after removing $\% 90$ of the dataset at random.
LargeVis and TriMap both
produce good results in that the clusters of the reduced
datasets are located roughly in the same arrangement as in the
full dataset. However, t-SNE moves the clusters around and
splits some of the original clusters into smaller spurious sub-clusters.

For labeled datasets, we might be interested in visualizing
a particular subset of the classes, leading to our second
test.
Figure~\ref{fig:tests}(2.1.b) 
gives the results on running the methods 
on just the subset of even digits of MNIST. 
As can be seen, only the TriMap method is able
to preserve the relative distances between all the clusters
after removing the odd digits, while the other methods place the remaining clusters at arbitrary positions.

\subsection{Outlier test} 
Natural datasets frequently contain outliers
due to measurement or experimental errors. 
Detecting and removing the outliers is key step in statistical analysis 
that should be facilitated with the DR methods.
We can check the ability of the DR methods to reveal 
outliers by adding artificial outlier points to the
original dataset. Figure~\ref{fig:tests}(2.2) shows the
DR results after shifting point 
\texttt{\bf X} 
(originally placed inside cluster `8' by all methods)
far away in a random direction. The point \texttt{\bf X}
is clearly an outlier in the new dataset.
Surprisingly, t-SNE and LargeVis both place the outlier
inside the cluster `0', which happens to be the
nearest cluster to this point in higher dimension. 
Even more disturbingly, adding a single outlier
this way consistently
rearranges the relative location of the clusters for t-SNE and LargeVis. 
On the other hand, TriMap successfully shows the outlier point far away from the rest of the dataset, while preserving the structure of the data.

\subsection{Multiple scales test} 
A DR tool should be able to reflect the global structure of
the data at different scales. For instance, 
a dataset might consist of multiple clusters where each cluster itself 
may contain multiple sub-clusters and so on. 
The practitioner can do a rudimentary test
by duplicating and shifting the natural dataset at hand. 
That is, for each point $x_n$ of
the original dataset, we add a point $x_n+c$, where $c$ is a fixed random
shift which is chosen large enough such that the two copies are guaranteed to be far apart in the high-dimensional space\footnote{%
Similar results obtained by shifting the datasets in 4
directions.}. When applied on the duplicated dataset, the DR method should be able to show each copy separately. Figure~\ref{fig:tests}(2.3) illustrates the
results on a duplicated MNIST dataset. We expect to see two
identical copies of the same dataset in the low-dimensional
embedding. Curiously enough, both t-SNE and LargeVis fail
to provide any meaningful global structure. In fact, both
methods tend to split some of the clusters into smaller
sub-clusters. On the other hand, TriMap successfully
recovers the two identical copies, which look similar to
the original dataset in Figure~\ref{fig:tests}($\star$).

\section{Sketch of the t-SNE and LargeVis}

The t-SNE method~\cite{tsne} is perhaps the most commonly used DR method in practice. The main idea of t-SNE is to reflect the pairwise similarities of the points by minimizing a KL divergence between the normalized pairwise similarities in the high-dimensional space and the low-dimensional embedding. The t-SNE method has $\mathcal{O}(N^2)$ time complexity for $N$ points. However, the complexity can be reduced
to $O(N\log N)$ by approximating the gradient using tree-based algorithms~\cite{bhtsne}. t-SNE is easy to apply to general datasets and usually produces nicely separated clusters. However, as we showed in the previous section, the results are sometimes extremely misleading: the whole dataset is simply collapsed into an ``orb'' and the outliers are shown as ``inliers''. The latter can be explained by fact that for an outlier point, the pairwise similarities to the points that are closest to the outlier point dominate the rest, causing the method to pull back the outlier point to the closest cluster.

LargeVis is a more recent method \cite{largevis} that also aims
to preserve the {\em pairwise} (dis)similarities of the points in the high-dimensional space in the low-dimensional embedding. To avoid the $\mathcal{O}(N^2)$ complexity of t-SNE, LargeVis uses a negative sampling approach~\cite{negative} to randomly select a subset of the dissimilar pairs. However, the weights of all the dissimilar pairs are set to a positive constant.  This causes the LargeVis to lose global information in the data, as we showed in the previous section. Overall, LargeVis forms well-separated clusters. However, the outlier points that are far away or in-between the clusters are pushed back into the closest clusters. This is a consequence of using a noisy distribution for negative sampling which tends to pick points from the denser regions with higher probability. For example, placing a single point which lies between multiple large clusters inside the nearest cluster increases the likelihood of the model because a single point has a very small probability of being selected as a dissimilar example by any of points that lie inside the large clusters.


\section{The new TriMap Method}

The main idea in developing the TriMap method is to preserve a higher-order of structure in the data. In other words, we aim to reflect the relative (dis)similarities of triplets of points, rather than pairs of points. Formally, a triplet is a constraint between three points $i$, $j$, and $k$, denoted as an ordered tuple  $(i,j,k)$, which indicates: ``point $i$ is more similar to point $j$ than point $k$''.

\subsection{Problem Formulation}

Let $\{x_n\}_{n=1}^N$ denote the high-dimensional
representation of $N$ points. Our goal is to find a
lower-dimensional representation $\{y_n\}_{n=1}^N$ for
these points, possibly in 2D or 3D. Let $p_{ij} \geq 0$ be
the pairwise similarity function between $x_i$ and $x_j$ in
high-dimension. Let $q_{ij} \geq 0$ denote the similarity
function in the lower-dimension between $y_i$ and
$y_j$.\footnote{For instance in t-SNE, the $p_{ij}$ and
$q_{ij}$ are the Gaussian and Student t-distributions, respectively.} We denote by $T$ be the set of all triplets $(i,j,k)$ where $p_{ij} > p_{ik}$, i.e.,  $T = \{(i,j,k):\, p_{ij} > p_{ik}\}$. Similar to~\cite{ste}, we can define the \emph{satisfaction probability} of the triplet $\triplet$ as 
\[
\text{Pr}_{ijk} = \frac{q_{ij}}{q_{ij} + q_{ik}} = \frac{1}{1 + \frac{q_{ik}}{q_{ij}}}\, .
\]
Notice that $\text{Pr}_{ijk} \rightarrow 1$ whenever $q_{ij} \gg q_{ik}$. The low-dimensional embedding can be calculated by minimizing the weighted sum of negative log-probabilities
\begin{align}
\min_{\{y_n\}} -\!\!\!\!\!\! \sum_{\triplet \in T}\!\!\!\! \w_{ijk} \log \text{Pr}_{ijk}
= +\!\!\!\!\!\! \sum_{\triplet \in T}\!\!\!\! \w_{ijk} \log (1 + \frac{q_{ik}}{q_{ij}})\label{eq:loss-ste}
\end{align}
where\footnote{Note that the loss in~\cite{ste} is a
special case of~\eqref{eq:loss-ste} with $\w_{ijk} = 1$ for
all $\triplet \in T$.} $\w_{ijk}$ is the weight of the
triplet $\triplet$.
By writing the objective function as in~\eqref{eq:loss-ste}, we can alternatively view the ratio $0 \leq \frac{q_{ik}}{q_{ij}}$ as the loss of the triplet $\triplet$, i.e, as $q_{ik}$ becomes smaller and $q_{ij}$ becomes larger, the loss approaches zero. 
The transformation $\log(1 + \cdot)$ of the loss reduces the effect of each triplet 
This transformation has been used before in the context of ranking to reduce the effect of individual losses on the overall ordering~\cite{robirank}. 

Typically not all the triplets can be satisfied when
mapping from high-dimension to lower-dimension, i.e., we
cannot have $q_{ij} > q_{ik}$, for all $(i,j,k) \in T$. The
issue arises because of the lower degree of freedom in the 
low-dimensional space. As a simple example, 
consider mapping uniformly distributed points on a two-dimensional circle to a one-dimensional line; regardless of the choice of embedding, the end points of line will always violate some triplet constraints. 
For a highly violated triplet $(i,j,k)$ with $q_{ik} \gg q_{ij}$, 
the triplet loss $\frac{q_{ik}}{q_{ij}}$ 
will become too large and dominate the loss
minimization over all triplets. 
Thus damping the effect of individual triplets is crucial 
whenever the triplets are sampled from a high-dimensional space.
We will show that the $\log$-transformation 
of the triplet losses \eqref{eq:loss-ste} 
is not sufficient to handle unsatisfied triplets. 
Instead, we exploit the properties of the generalized
$\log$ and $\exp$ functions to create stronger damping functions.

\begin{figure}[t!]
\vspace{-2mm}
\begin{center}	
	\subfigure[]{\includegraphics[width=0.23\textwidth]{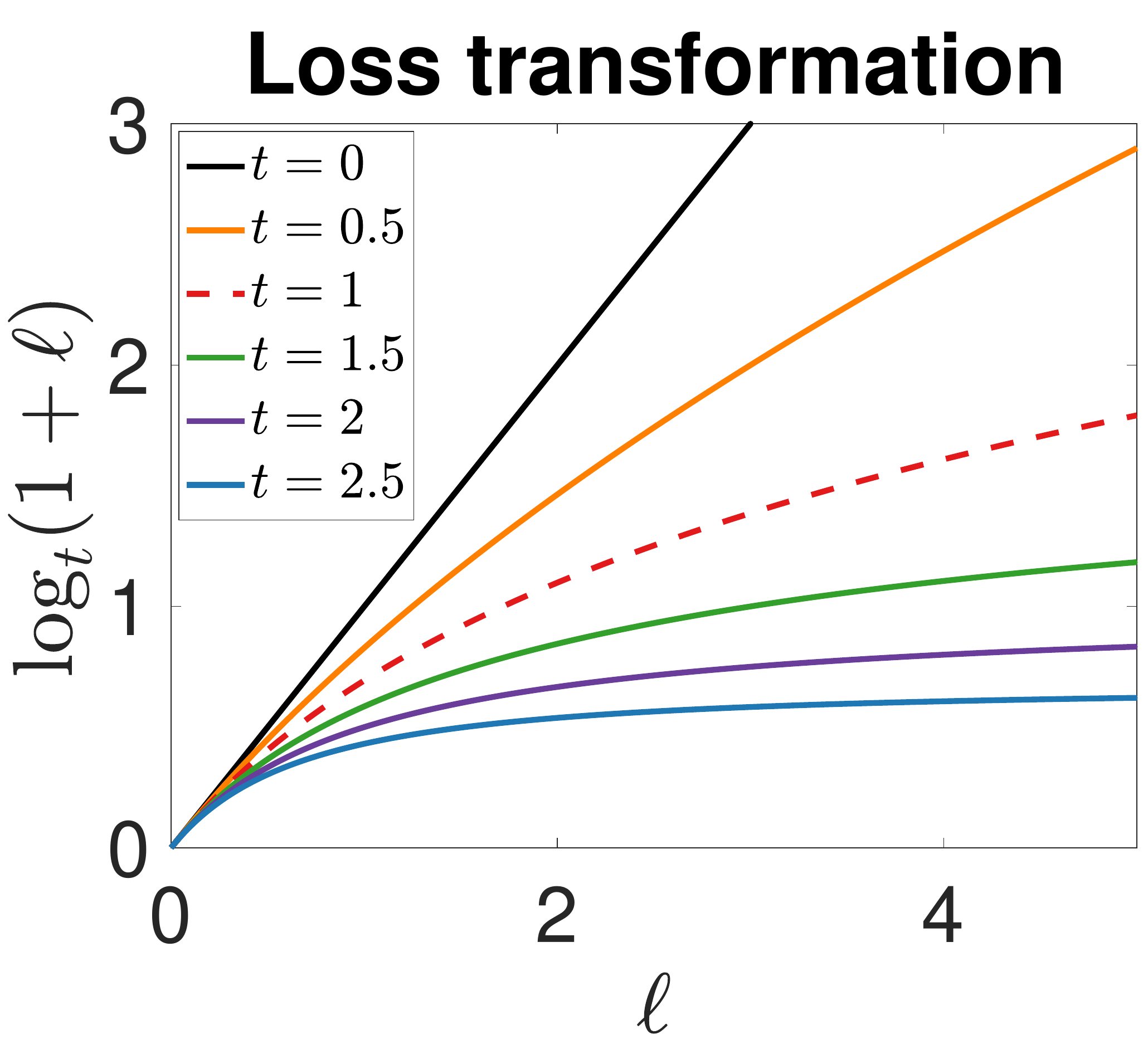}\label{fig:logt}}\,\,
     \subfigure[]{\includegraphics[width=0.23\textwidth]{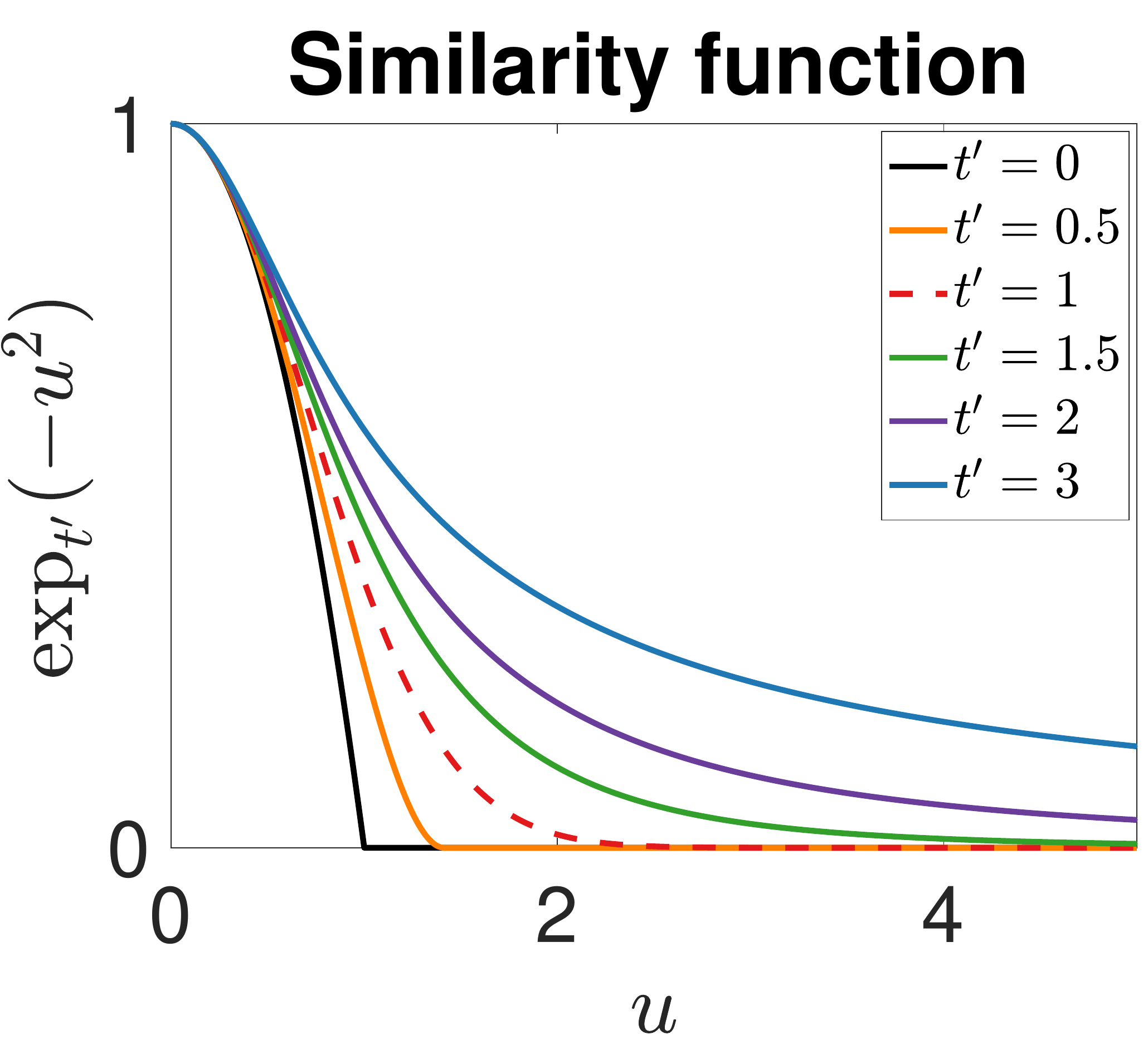}\label{fig:expt}}
     \caption{(a) Loss transformation $\ell \rightarrow \log_t(1 + \ell)$ and (b) similarity function $u \rightarrow \exp_{t^\prime}(-u^2)$ for different values of $t$ and $t^\prime$, respectively.}\label{fig:texp}
     \vspace{-2mm}
     \end{center}
\end{figure}

We propose the following generalized $\log_t$-transformation
\begin{equation}
\label{eq:loss-prim}
\min_{\{y_n\}} \sum_{\triplet \in T} \w_{ijk}\, \log_t (1 + \frac{q_{ik}^{(t^\prime)}}{q_{ij}^{(t^\prime)}})\, ,
\end{equation}
where, the $\log_t$ function~\citep{texp1} is defined as
\begin{equation*}
\label{eq:logt}
\log_t(x) = 
\begin{cases}
\log(x) & \text{if } t = 1\\
(x^{1-t}-1)/(1-t) & \text{otherwise}
\end{cases}\, .
\end{equation*}

Note that $\log_t$ is concave and non-decreasing and
generalizes the $\log$ function which is recovered at $t = 1$. Thus,~\eqref{eq:loss-prim}
includes the $\log$-transformation as a special case (see
Figure~\ref{fig:logt}). However, for $x > 1$, the $\log_t$
function with $t > 1$ grows slower than the $\log$ function
and reaches the value $1/(t-1)$ in the limit $x \rightarrow
\infty$. Therefore, for $t > 1$, each triplet can
contribute at most $1/(t-1)$ to the total loss 
and the objective function of TriMap \eqref{eq:loss-prim} is robust to violated
triplets.

Furthermore, inspired by the well-behaved properties of the
heavy-tailed distributions for low-dimensional
embedding~\cite{tsne, hssne}, we define a parameterized
similarity function 
\begin{equation*}
\label{eq:q}
q_{ij}^{(t^\prime)} = \exp_{t^\prime}(-\Vert y_i - y_j\Vert^2),\,\, \text{ where,}
\end{equation*}
\begin{equation*}
\label{eq:expt}
\exp_{t^\prime}(x) = 
\begin{cases}
\exp(x) & \text{if } {t^\prime} = 1\\
[1 + (1-{t^\prime}) x]_+^{1/(1-{t^\prime})} & \text{otherwise}
\end{cases}\, ,
\end{equation*}
where $[\,\cdot\,]_+ = \max(0,\cdot)$. Notice that the
tail-heaviness of $q_{ij}^{(t^\prime)}$ increases with the
value of $t^\prime$. That is, the Gaussian function is
recovered at $t^\prime = 1$ and
$t^\prime = 2$ amounts to the Student t-distribution with
one degree of freedom. For $t^\prime > 2$, we recover
distributions with even heavier tails. 
Figure~\ref{fig:expt} illustrates the $q_{ij}^{(t^\prime)}$ function 
for several values of ${t^\prime}$. The default parameters for TriMap are set to
$t = t^\prime = 2$, but we will briefly explore the other choices in Section~\ref{sub:t}.

Another crucial component of TriMap is the weight
$\w_{ijk}$ assigned to the triplet $\triplet$, 
enforcing the extent to which $\triplet$ needs to be satisfied. We use $\w_{ijk} = \frac{p_{ij}}{p_{ik}}$ where
$$
p_{ij} = \exp(-\frac{\Vert x_i - x_j\Vert^2}{\sigma^2_{ij}})\, ,
$$
and $\sigma^2_{ij} = \sigma_i\, \sigma_j$. The scaling factor $\sigma_i$ is set to the average distance of $x_i$ to its $10$-th to $20$-th nearest-neighbors. This choice of $\sigma_i$ adaptively  adjusts the scale of $p_{ij}$ based on the density of the data~\cite{zelnik}.

\begin{figure*}[th!]
\vspace{-2mm}
\begin{center}
\subfigure{\includegraphics[width=0.19\textwidth]{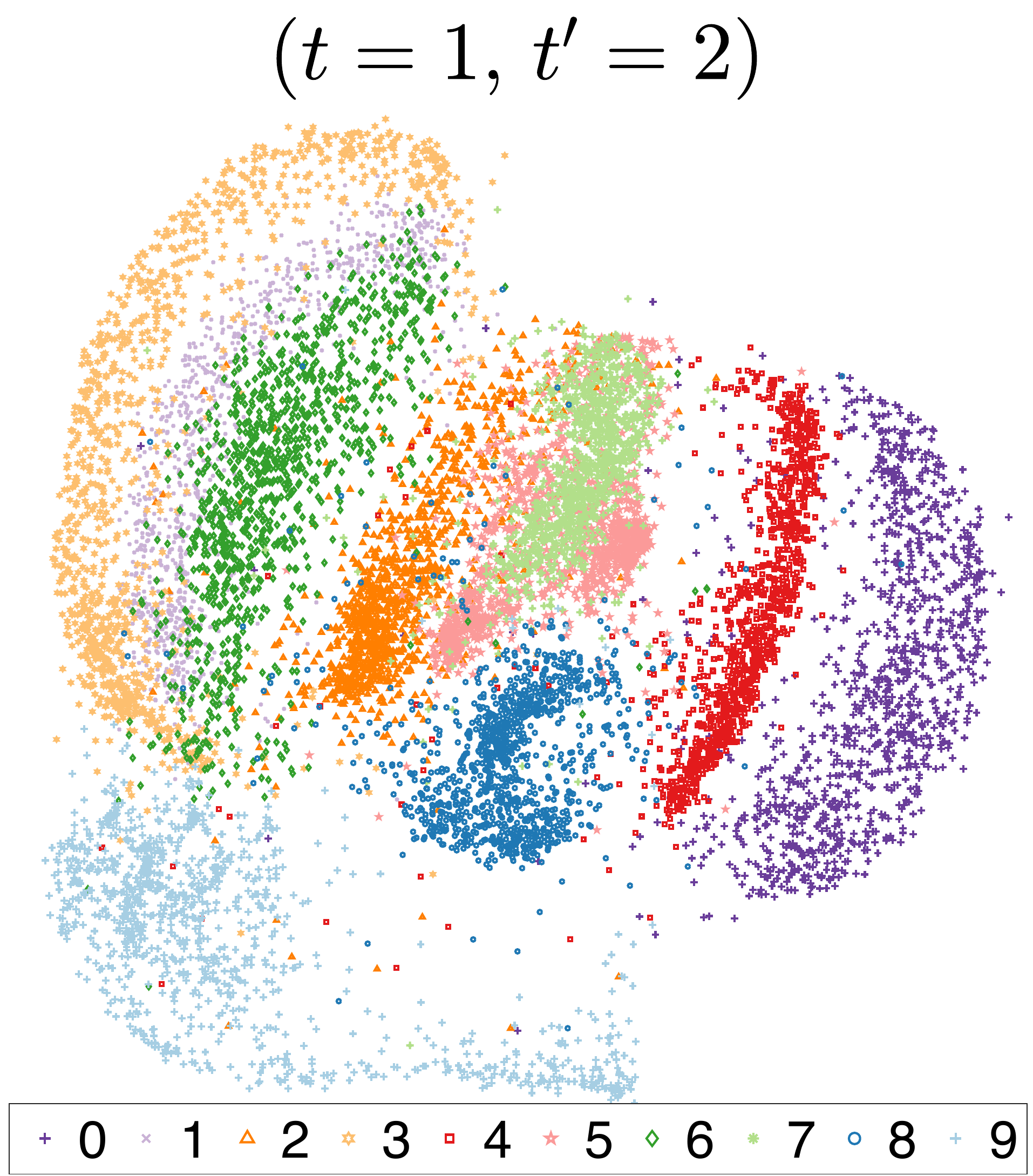}}
\subfigure{\includegraphics[width=0.19\textwidth]{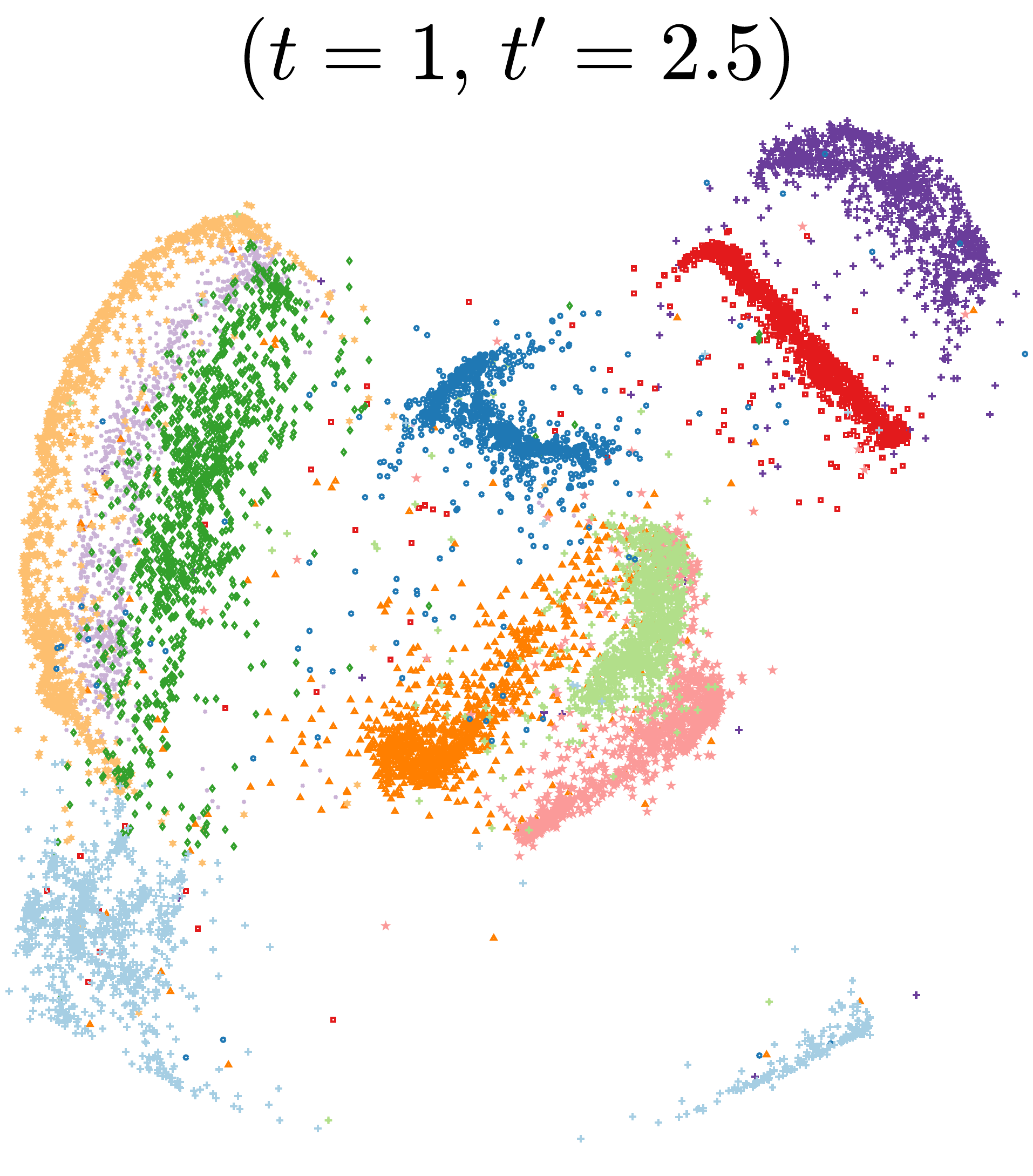}} 
\subfigure{\includegraphics[width=0.19\textwidth]{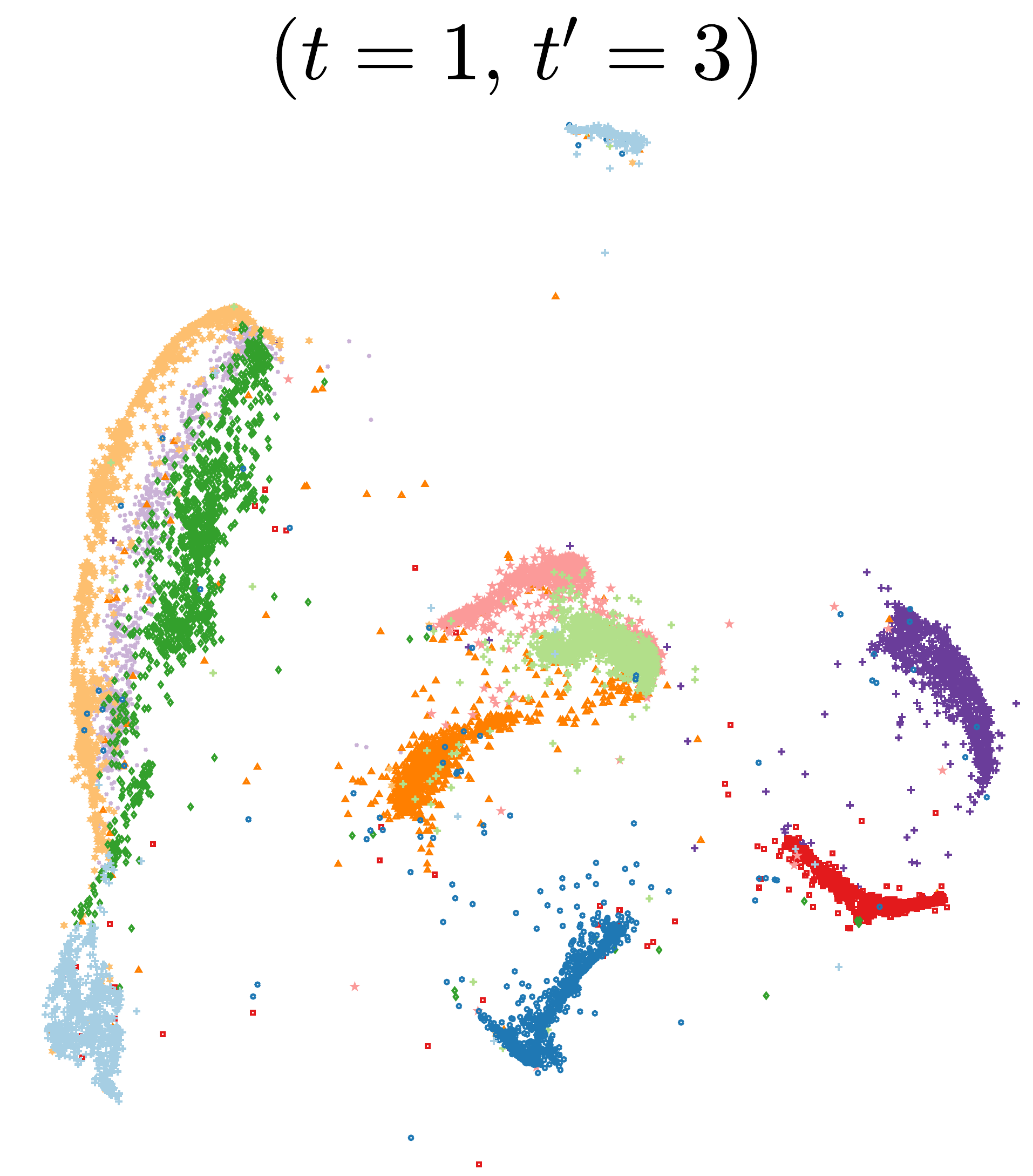}} 
\subfigure{\includegraphics[width=0.19\textwidth]{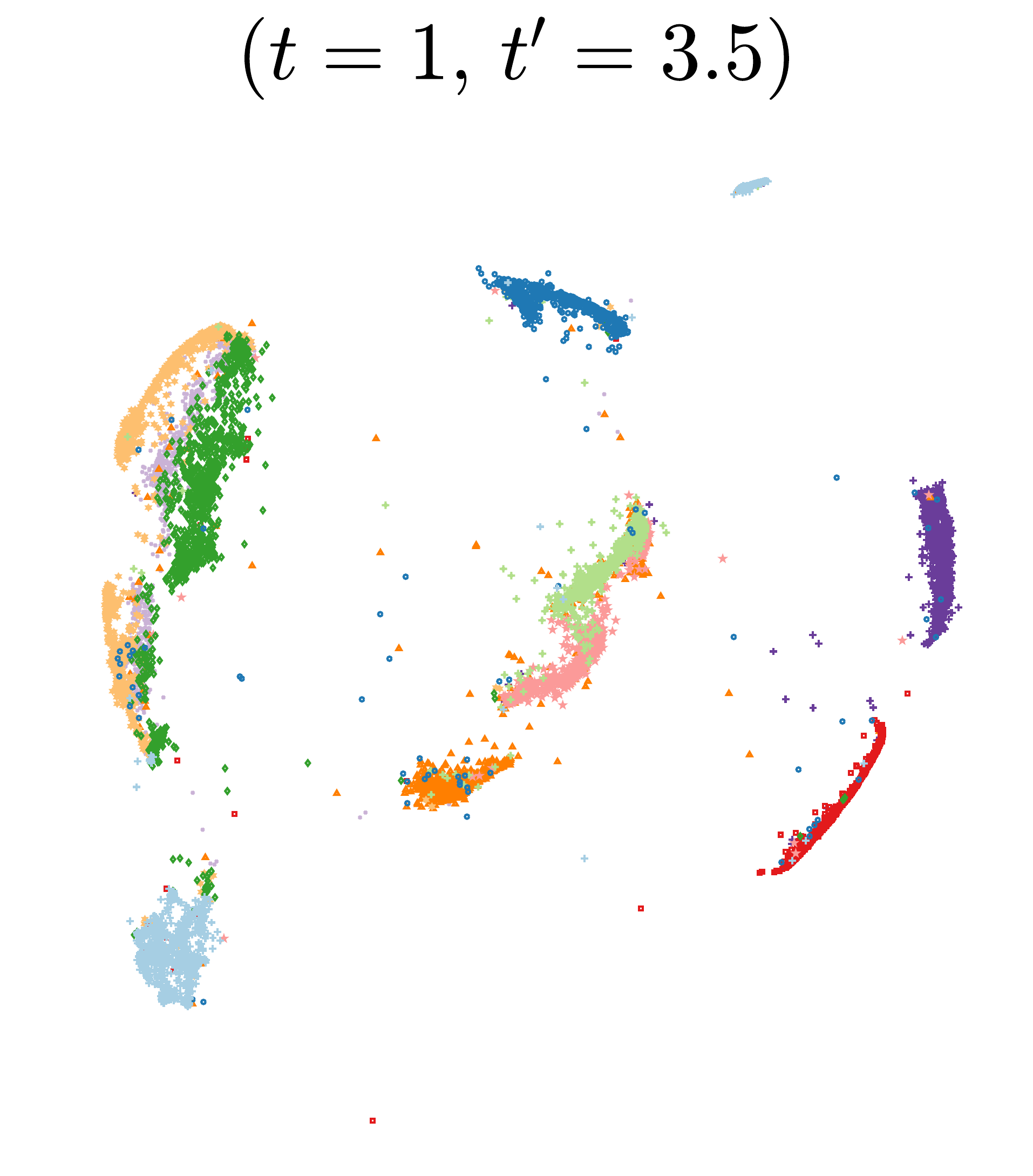}} 
\subfigure{\includegraphics[width=0.19\textwidth]{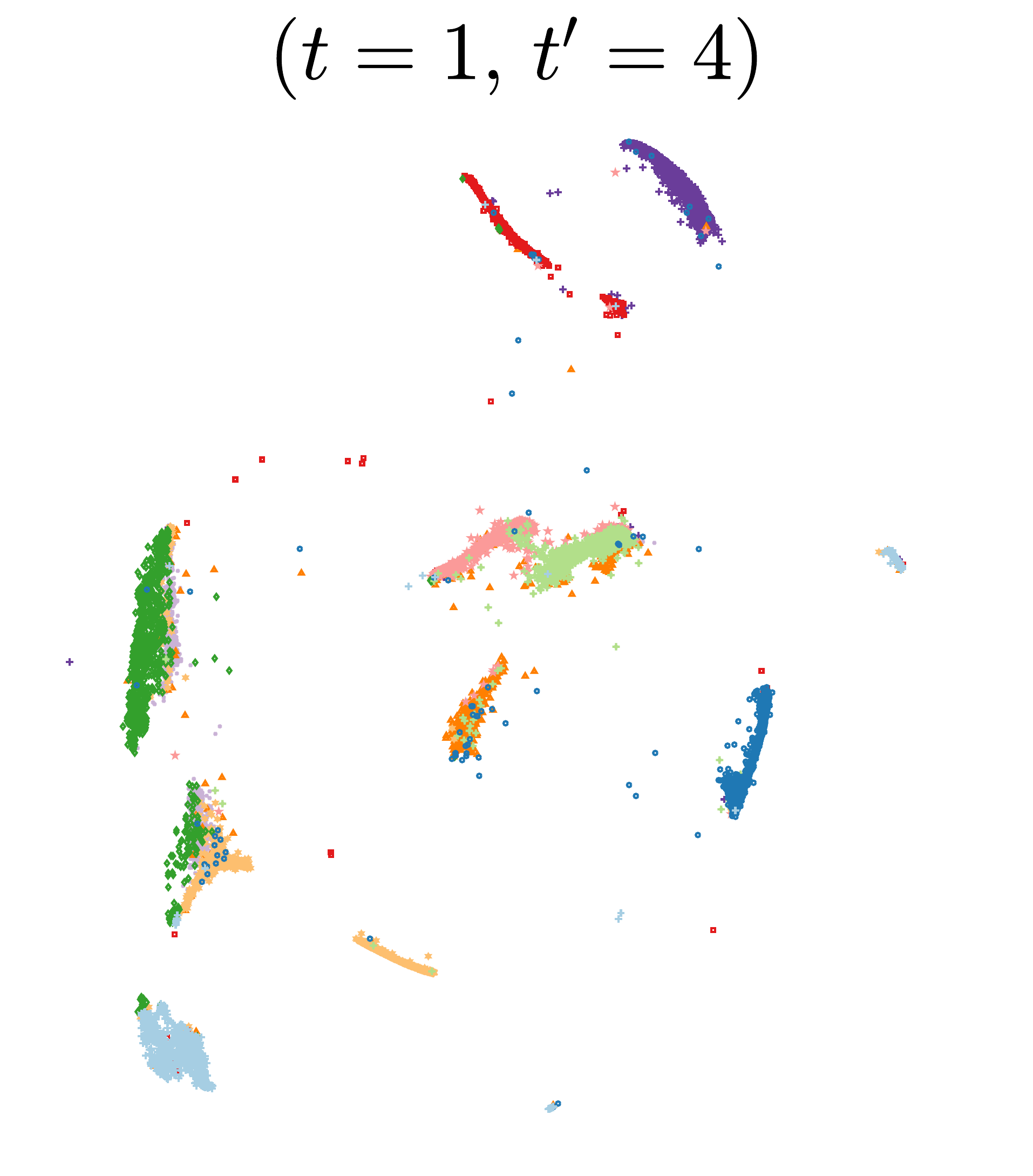}}\hfill\\
\subfigure{\includegraphics[width=0.19\textwidth]{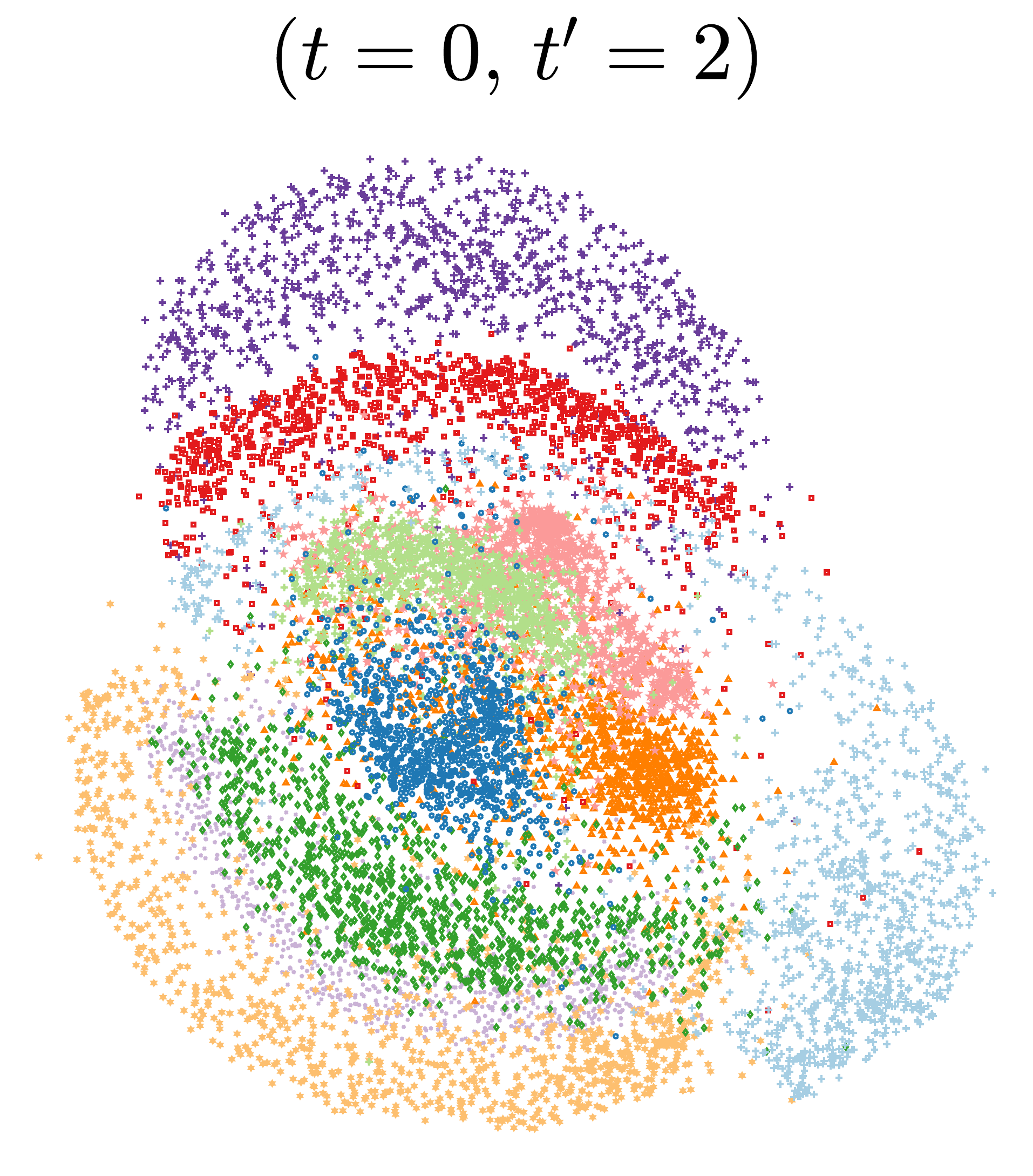}}
\subfigure{\includegraphics[width=0.19\textwidth]{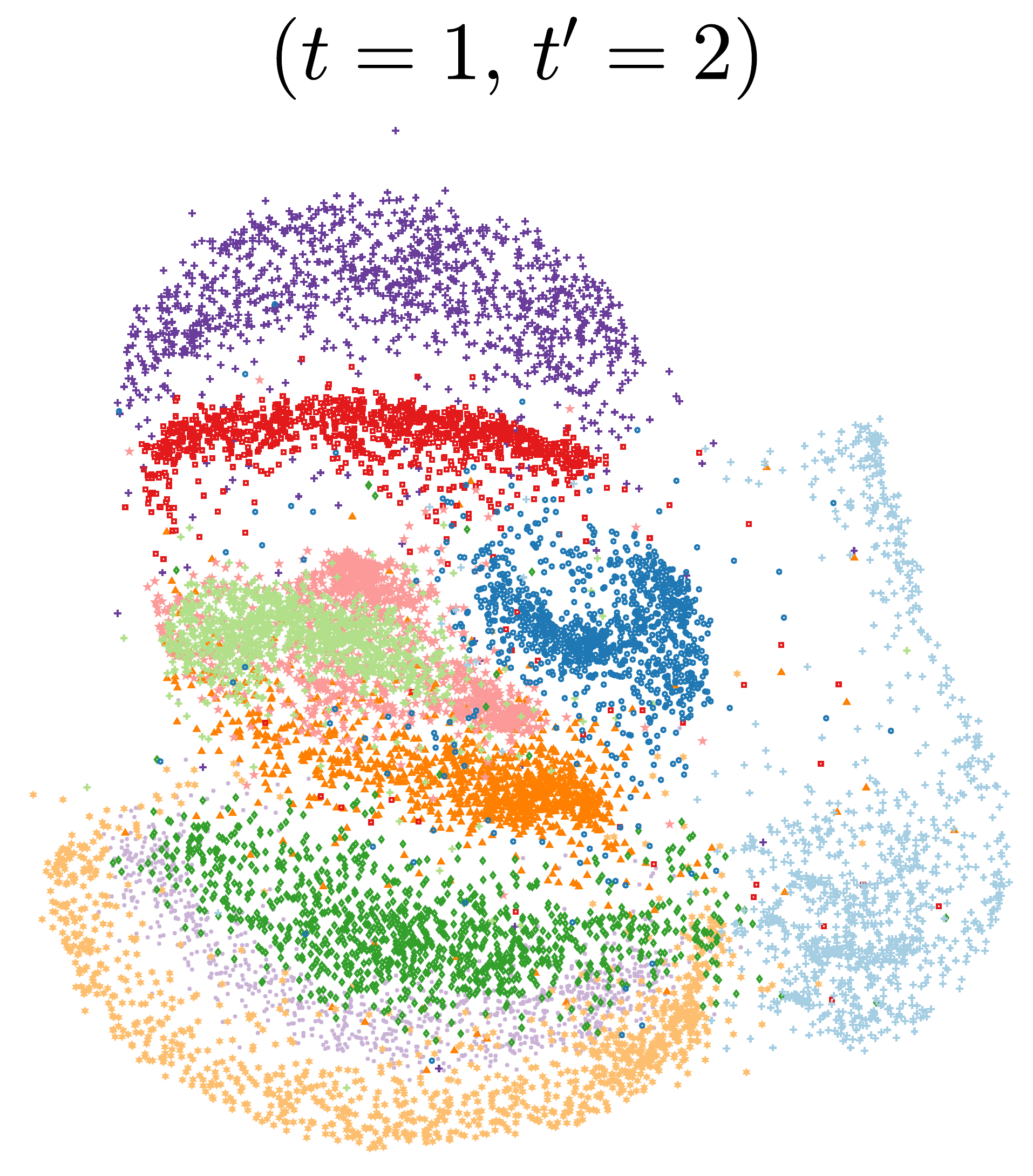}}
\subfigure{\frame{\includegraphics[width=0.19\textwidth]{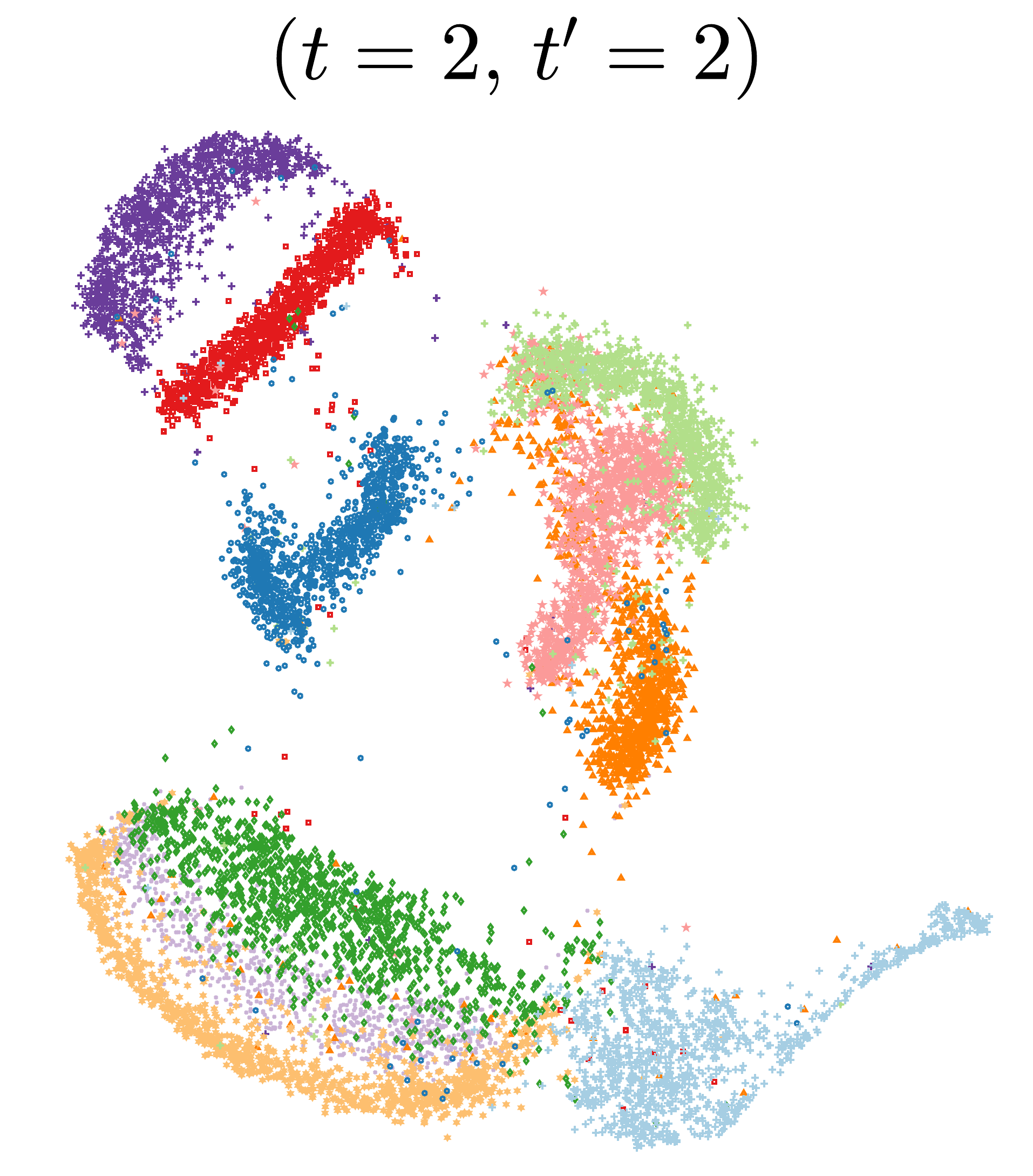}}} 
\subfigure{\frame{\includegraphics[width=0.19\textwidth]{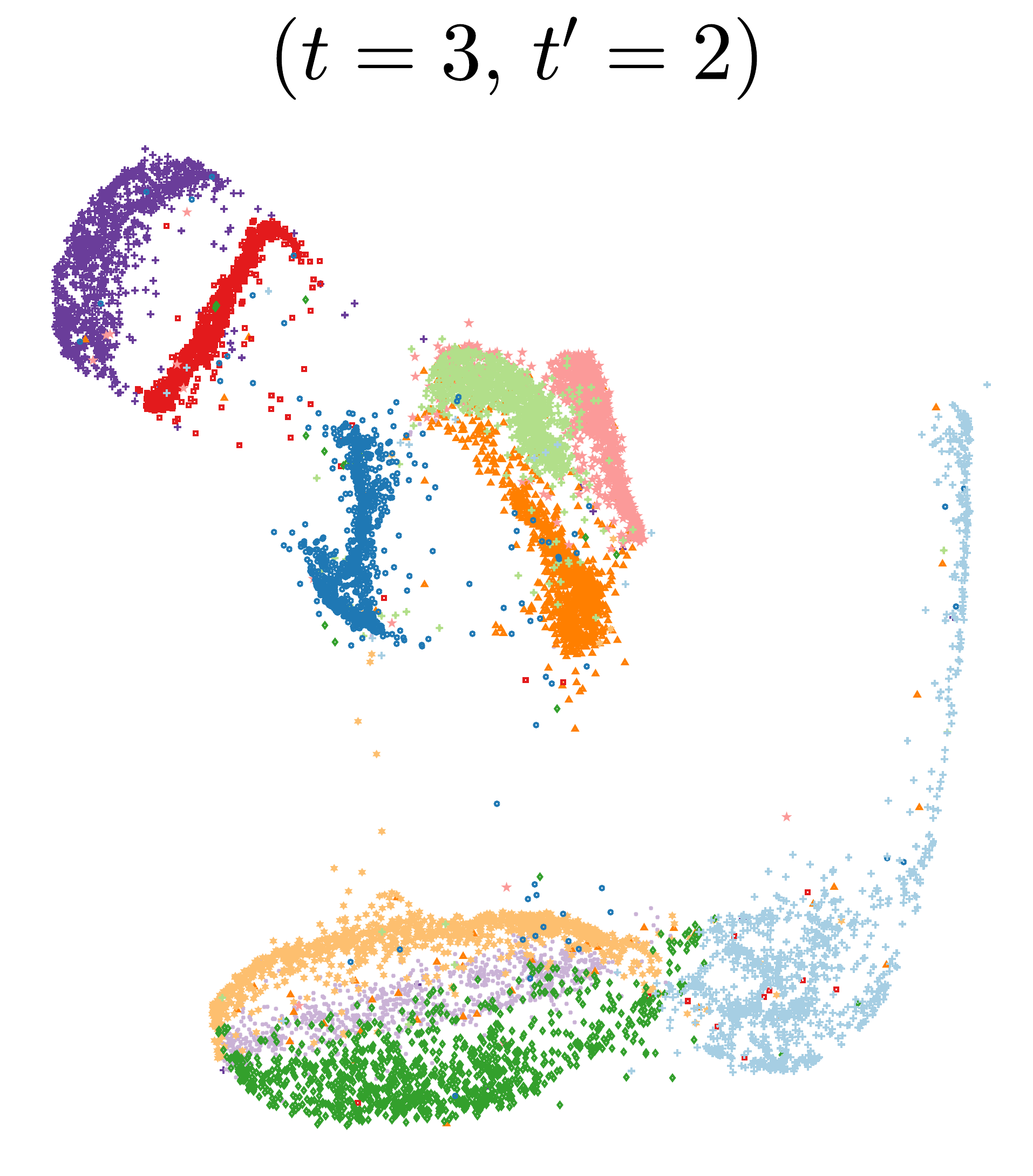}}} 
\subfigure{\frame{\includegraphics[width=0.19\textwidth]{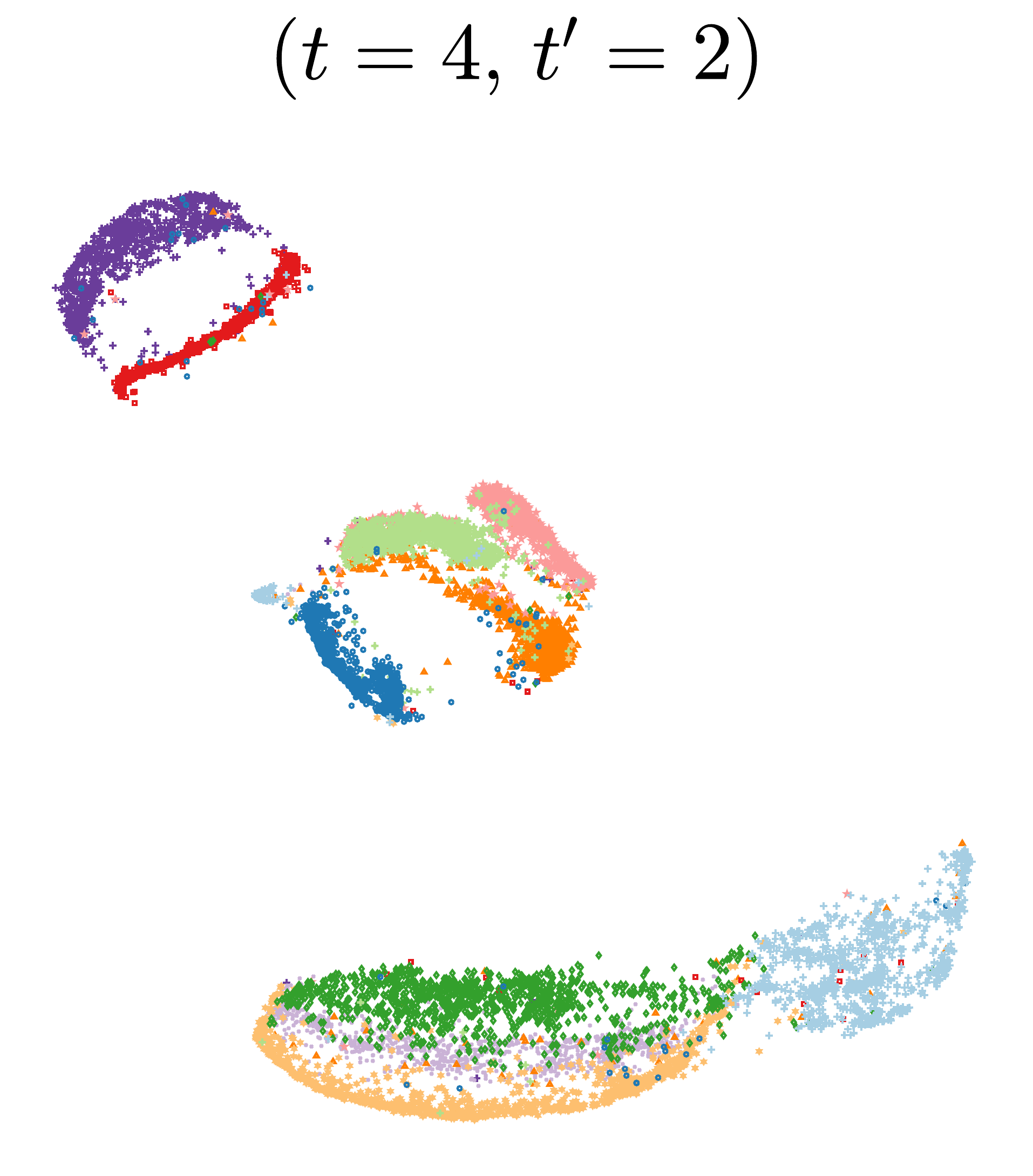}}}\hfill
\end{center}
\caption{\label{fig:mit-t1} Effect of changing $t$ and $t^\prime$ parameters: top) results using fixed $\log$-transformation ($t=1$) and different values of $t^\prime$ for similarity function. Notice that having heavier-tail provides more separability but cannot fix the clutter introduced by the unsatisfied triplets by itself, bottom) results using different values of $t$ (i.e. $\log_t$-loss transformation) with $t^\prime = 2$ (i.e. Student t-distribution for similarities) fixed. Note that having $\log_t$-transformation is crucial to obtain nice visualizations. Value of $t=0$ corresponds to no transformation (linear loss), $t=1$ recovers the $\log$-transformation, $t=2$ is used in our experiments. The boxes indicate the cases with acceptable choices of parameters by our algorithm.}
\vspace{-0.6cm}
\end{figure*}

\subsection{Triplet Sampling}

The direct minimization of~\eqref{eq:loss-prim} using all the possible triplets can be
too expensive for large datasets since the total
number of triplets grows as $\mathcal{O}(N^3)$ with number of
data points $N$. However, in practice, there exists a large
amount of redundancy among the triplets. For instance, the
two triplets $(i,j,k)$ and $(i,j,k^\prime)$, in which $k$
and $k^\prime$ are located close together, convey the same
information. We now develop a heuristic for sampling
a linear size subset of triplets, such that the mapping produced
from the subset essentially reproduces the mapping derived from all
triplets (evidence not shown for lack of space). We sample two types of triplets to form the embedding.

\textbf{Nearest-neighbors triplets:} 
For each point $i$, the closer point $j$ is selected from $m$-nearest neighbors of the point and the distant points are sampled uniformly from the subset of points that are farther away from $i$ than $j$. Note that this choice of triplets reflect the local as well as the global information of the data. For each point in the set of $m$-nearest neighbors, we sample $m^\prime$ distant points which results in $N\times m\times m^\prime$ triplets in total. Note that in practice $m, m^\prime \ll N$ and we still have $\mathcal{O}(N)$ complexity. In all our experiments, we use $m=50$ and $m^\prime = 10$.

\textbf{Random triplets:} These triplets are sampled uniformly at random 
from the set of full triplet set $T$. 
Random triplets help preserve the global structure of the data. We add $s$ randomly selected triplets for each points, which results in $N\times s$ triplets in total. In all our experiments, we set $s = 5$.

In practice, dividing each ratio $\frac{p_{ij}}{p_{ik}}$ at the end by the maximum ratio among the triplets in the set of sampled triplets and adding a constant positive bias $\gamma > 0$ to all ratios improves the results. In all our experiments, we set $\gamma = 0.001$. 

\begin{figure*}[th!]
\vspace{-2mm}
\begin{center}
\begin{tabular}{m{0.25\textwidth} m{0.25\textwidth} m{0.25\textwidth}}
\multicolumn{1}{c}{t-SNE} & \multicolumn{1}{c}{LargeVis} &  \multicolumn{1}{c}{TriMap}\\
\includegraphics[width=0.22\textwidth]{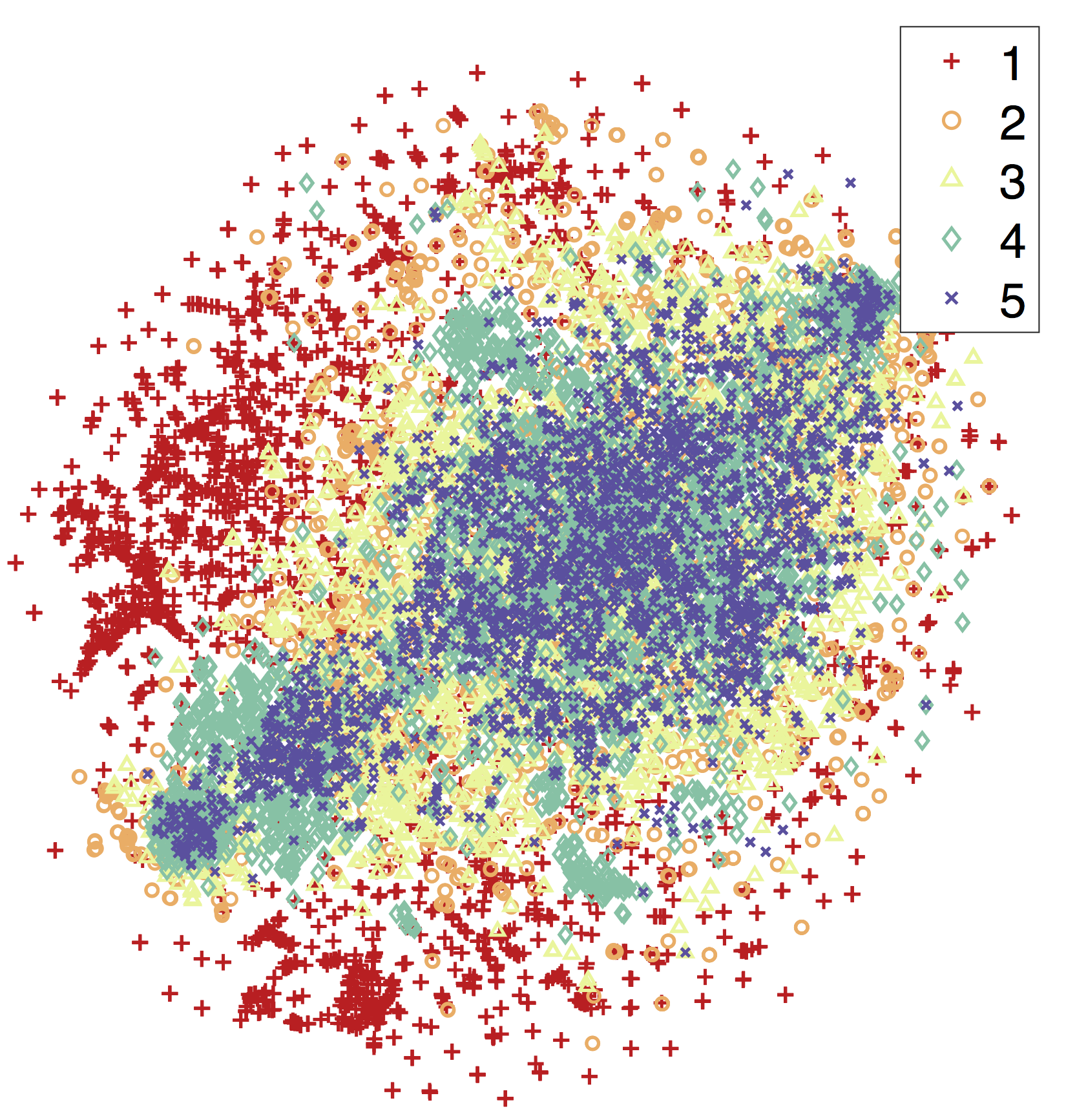} & \includegraphics[width=0.22\textwidth]{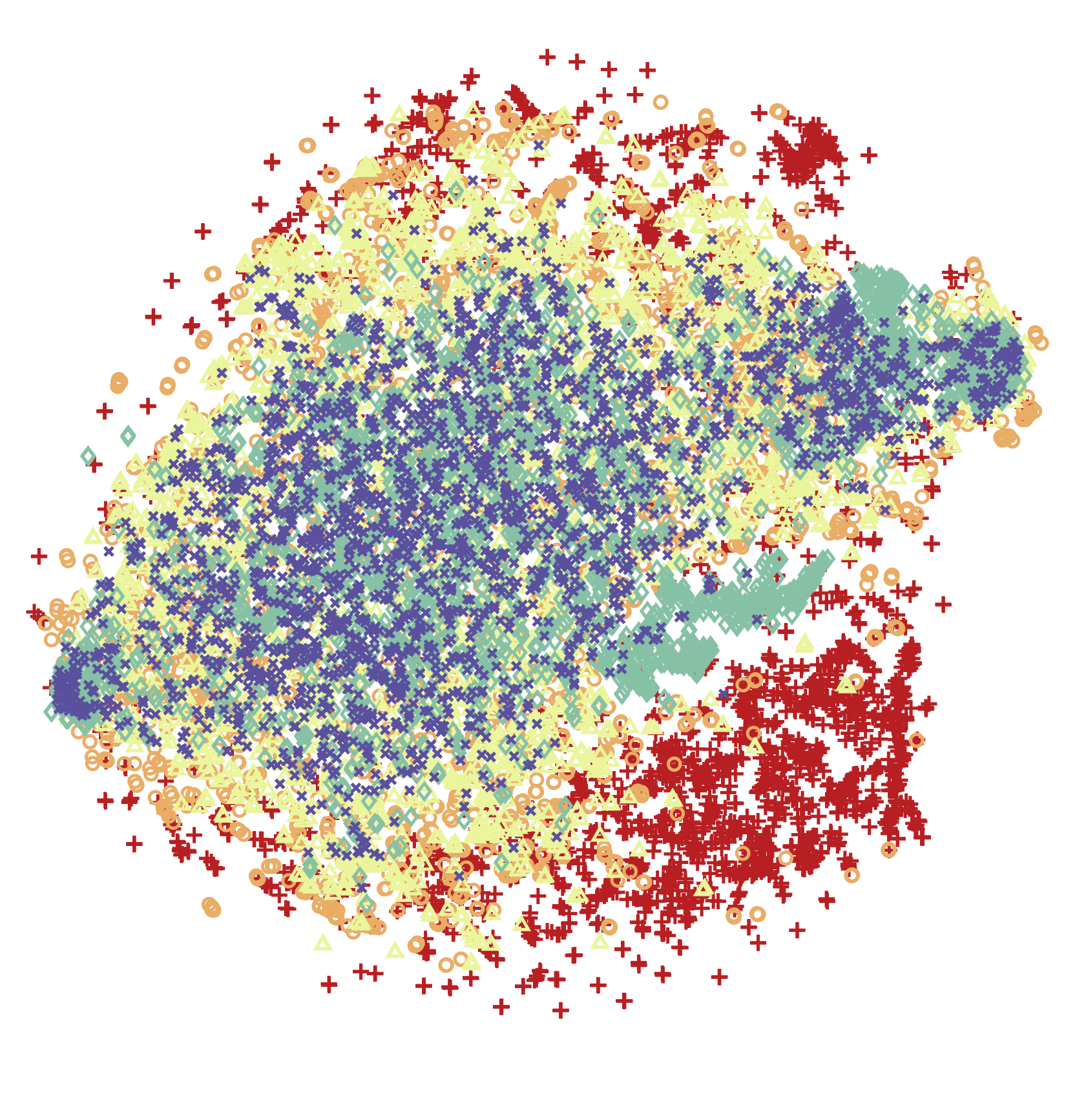} &   \includegraphics[width=0.22\textwidth]{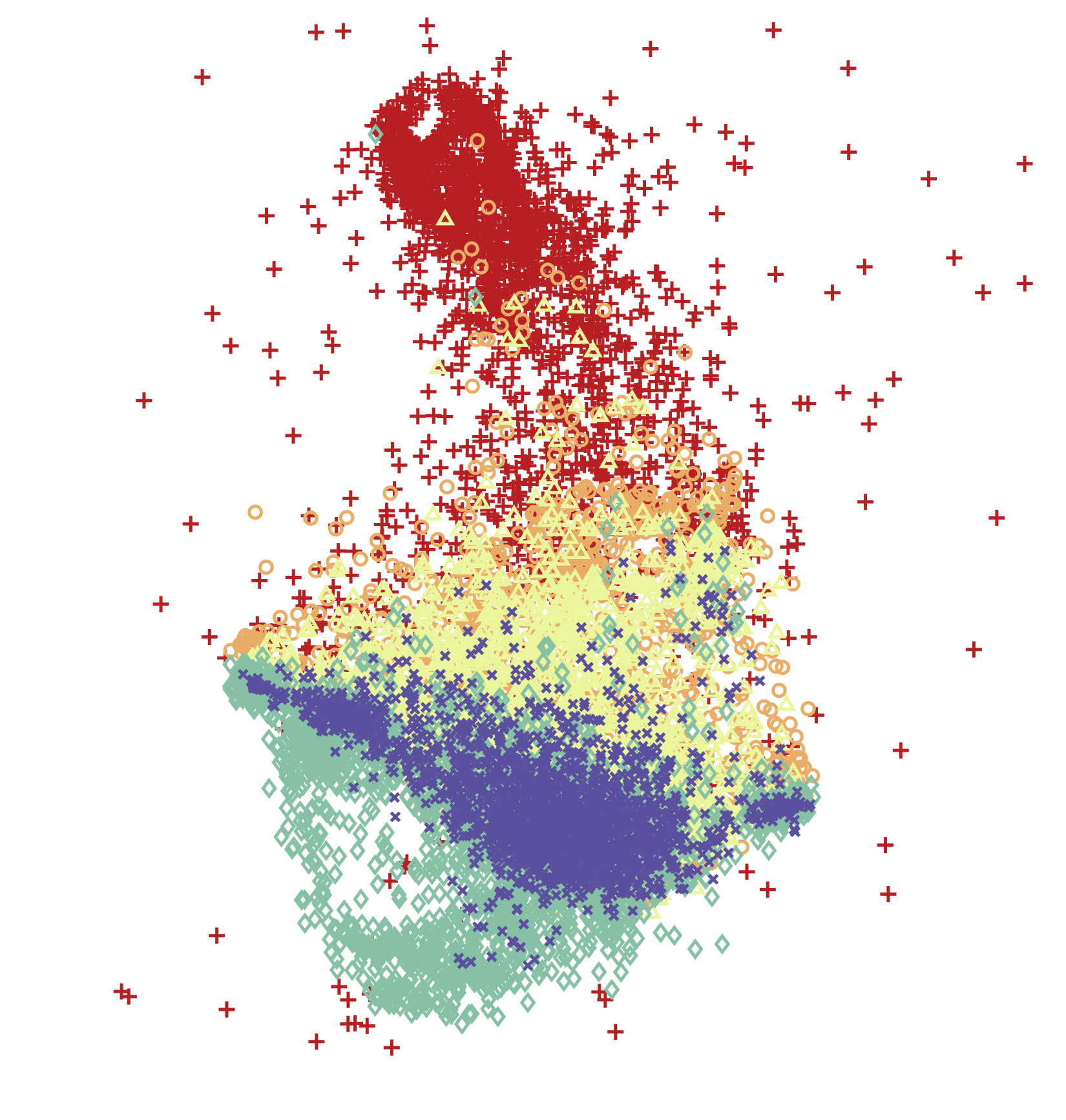}\hfill\\
\end{tabular}
\caption{\label{fig:epileptic} Epileptic Seizure Recognition dataset. TriMap shows a smooth transition between measurements under different conditions while both t-SNE and LargeVis fail to preserve the global structure.}
\end{center}
\vspace{-0.5cm}
\end{figure*}


\begin{figure*}[th!]
\begin{center}
\begin{tabular}{m{0.25\textwidth} m{0.25\textwidth} m{0.25\textwidth}}
\multicolumn{1}{c}{t-SNE} & \multicolumn{1}{c}{LargeVis} &  \multicolumn{1}{c}{TriMap}\\
\includegraphics[width=0.25\textwidth]{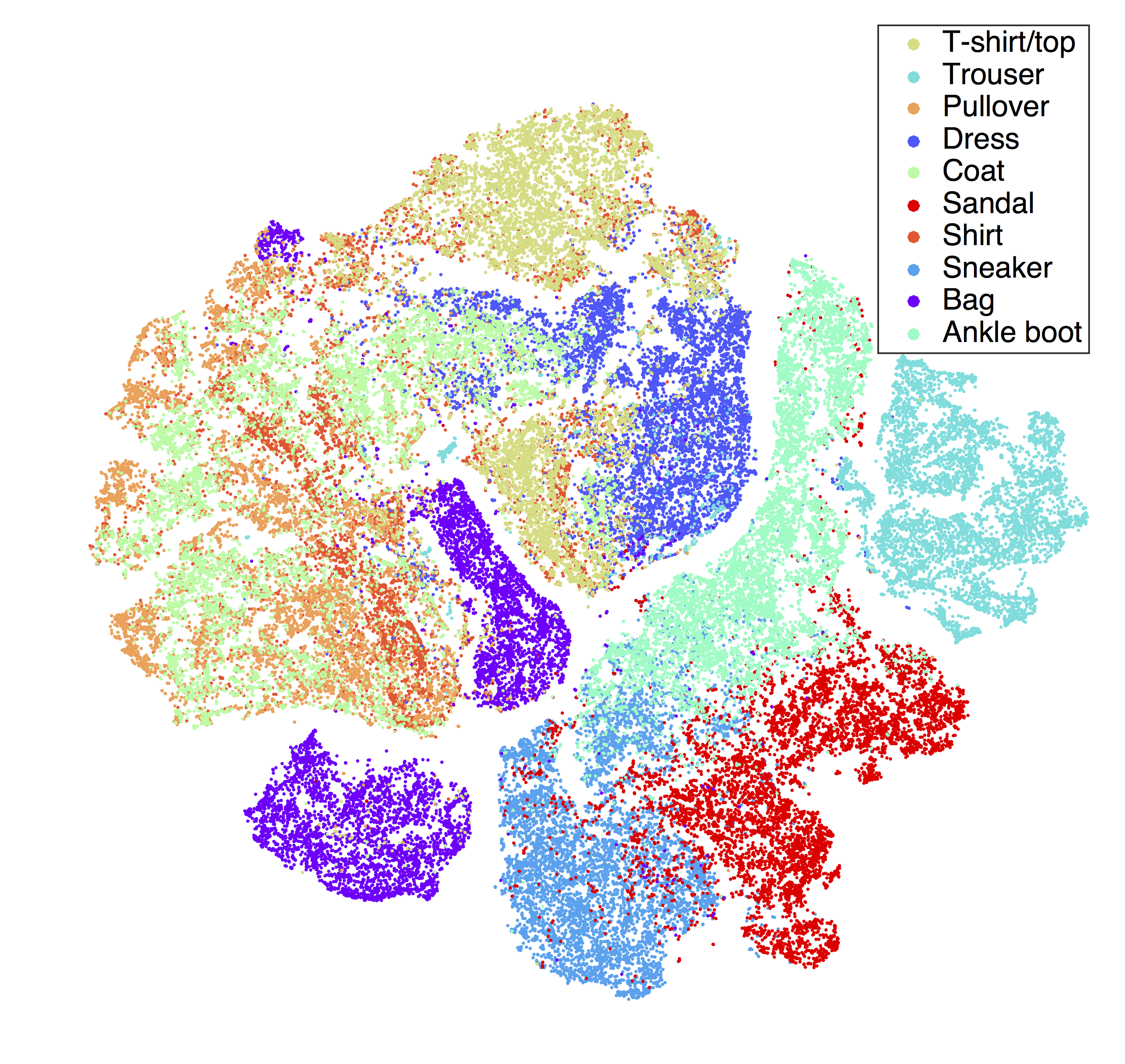} & \includegraphics[width=0.25\textwidth]{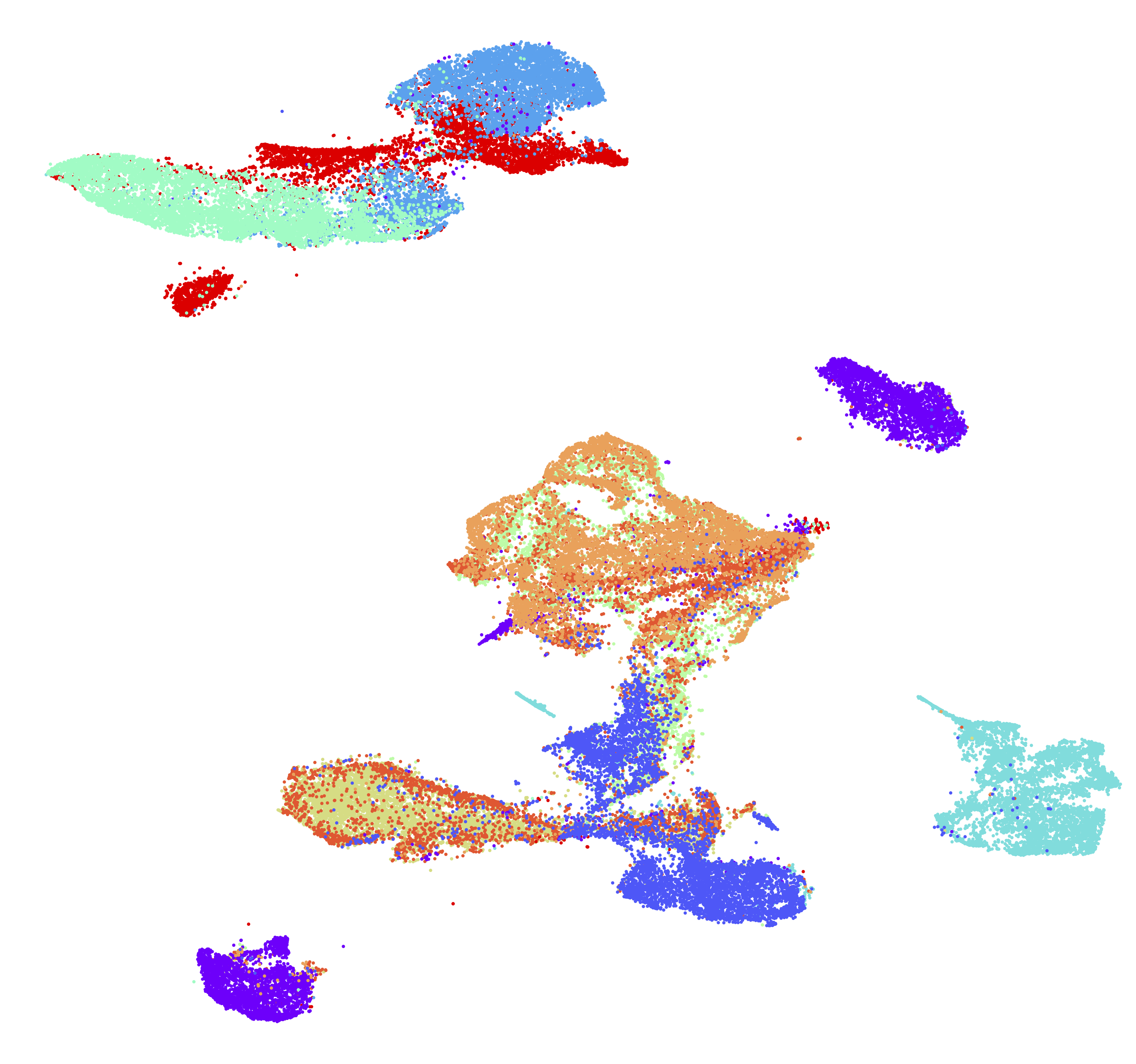} &   \includegraphics[width=0.25\textwidth]{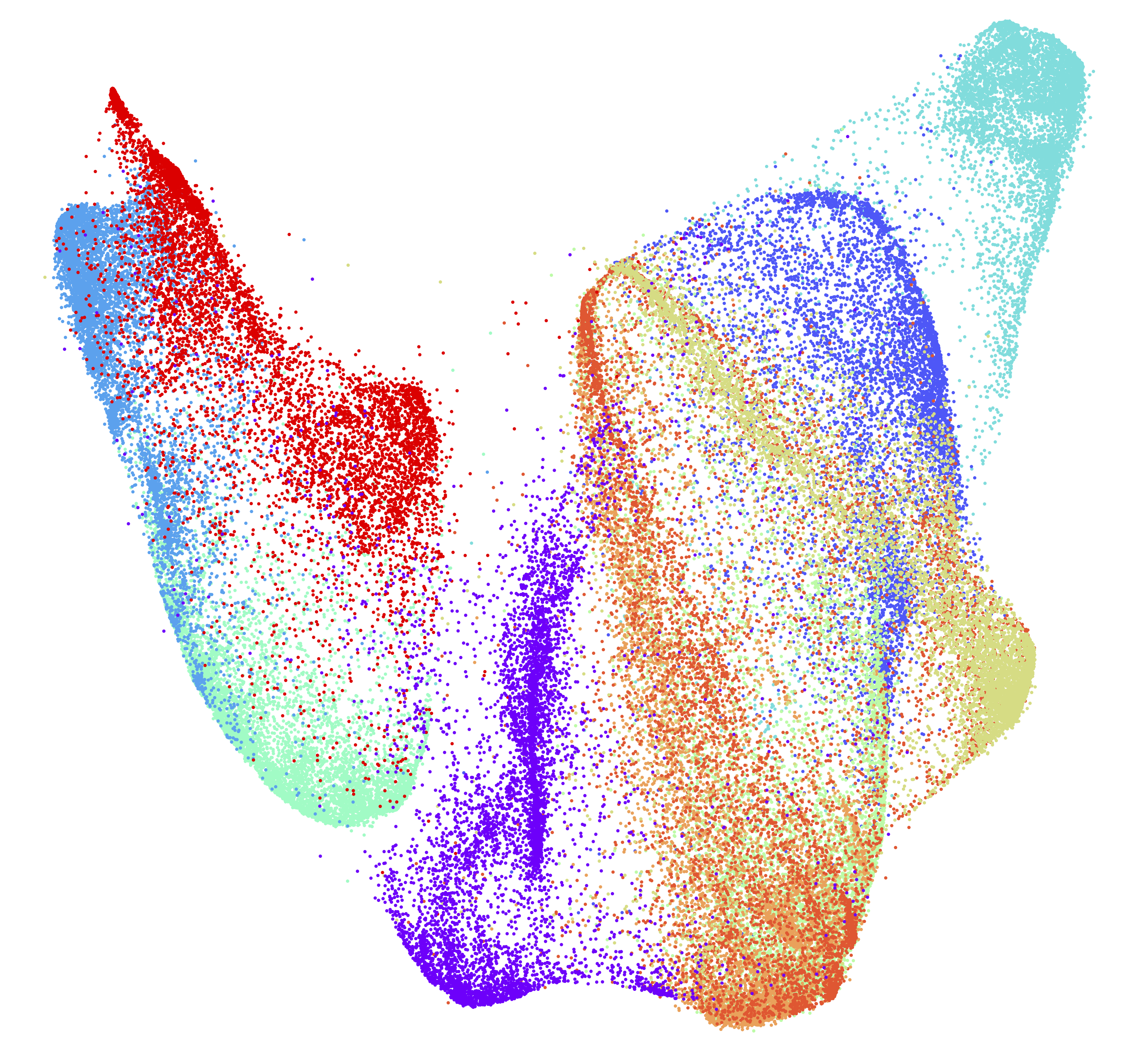}\hfill\\
\end{tabular}
\caption{\label{fig:fmnist} Fashion MNIST dataset: TriMap represents a smooth embedding where each class in concentrated around the same region and there appears to be a smooth transitions among different classes.}
\end{center}
\vspace{-0.5cm}
\end{figure*}

\begin{figure*}[th!]
\begin{center}
\begin{tabular}{m{0.25\textwidth} m{0.25\textwidth} m{0.25\textwidth}}
\multicolumn{1}{c}{t-SNE} & \multicolumn{1}{c}{LargeVis} &  \multicolumn{1}{c}{TriMap}\\
\includegraphics[width=0.25\textwidth]{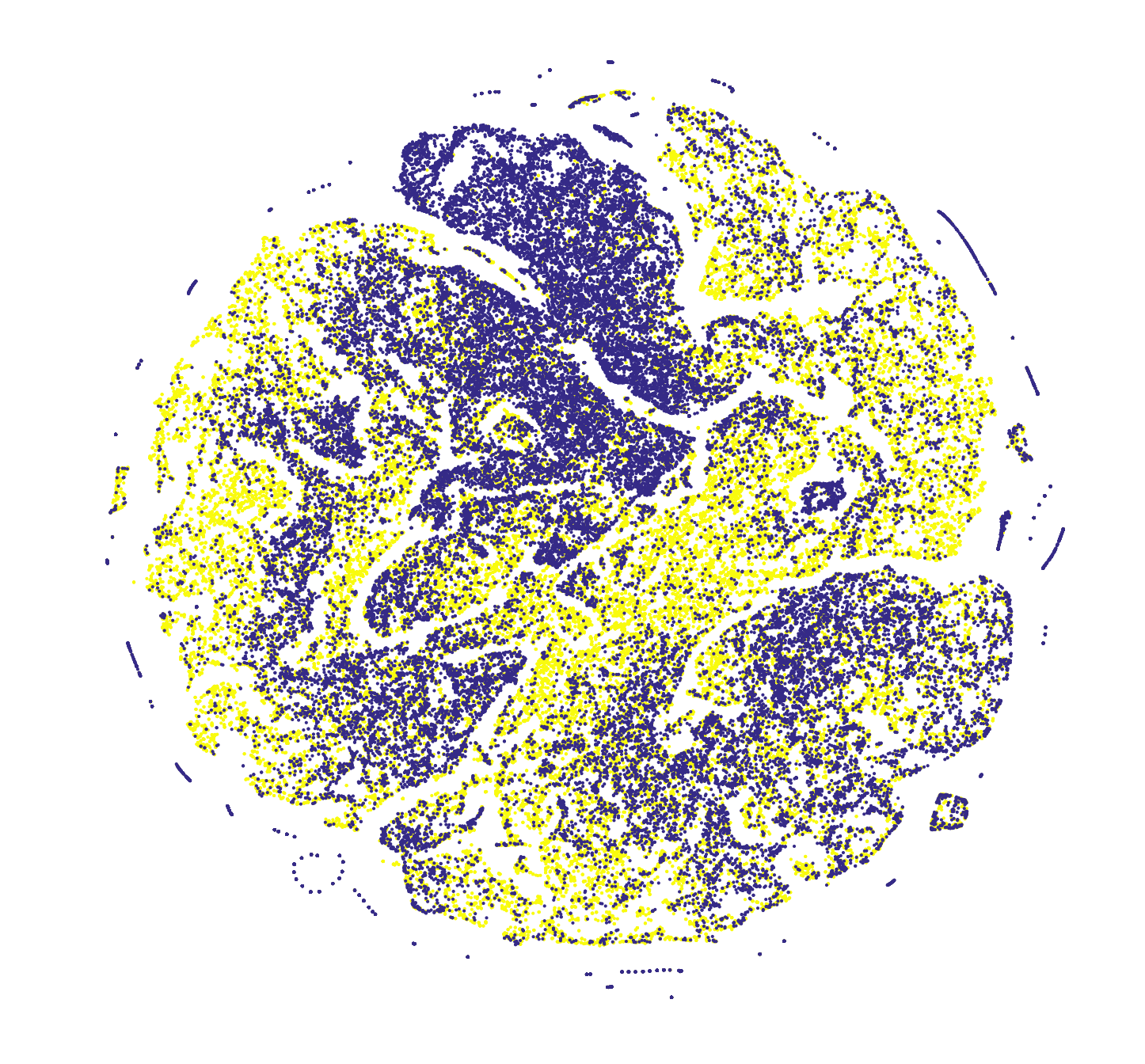} & \includegraphics[width=0.25\textwidth]{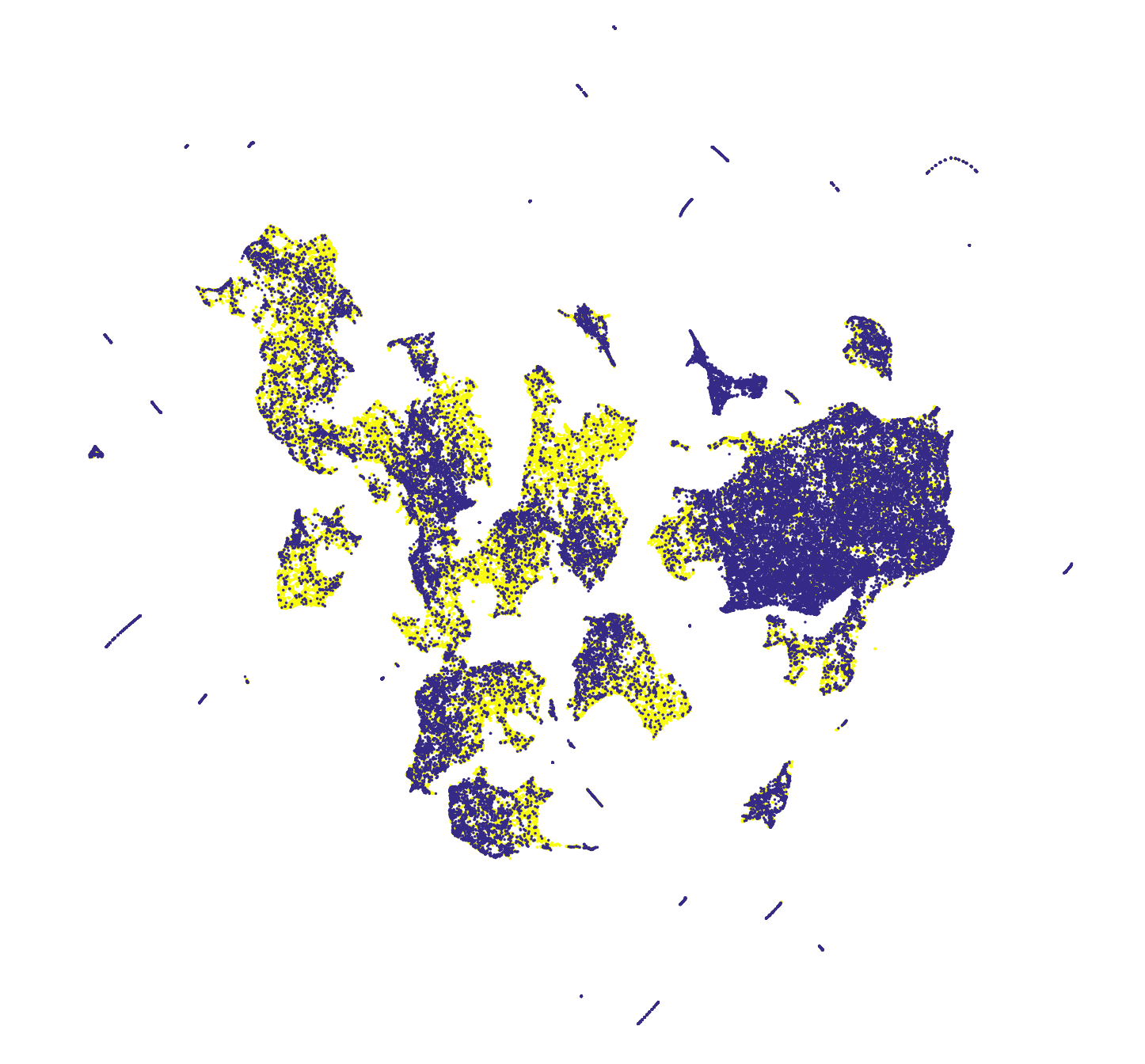} &   \includegraphics[width=0.25\textwidth]{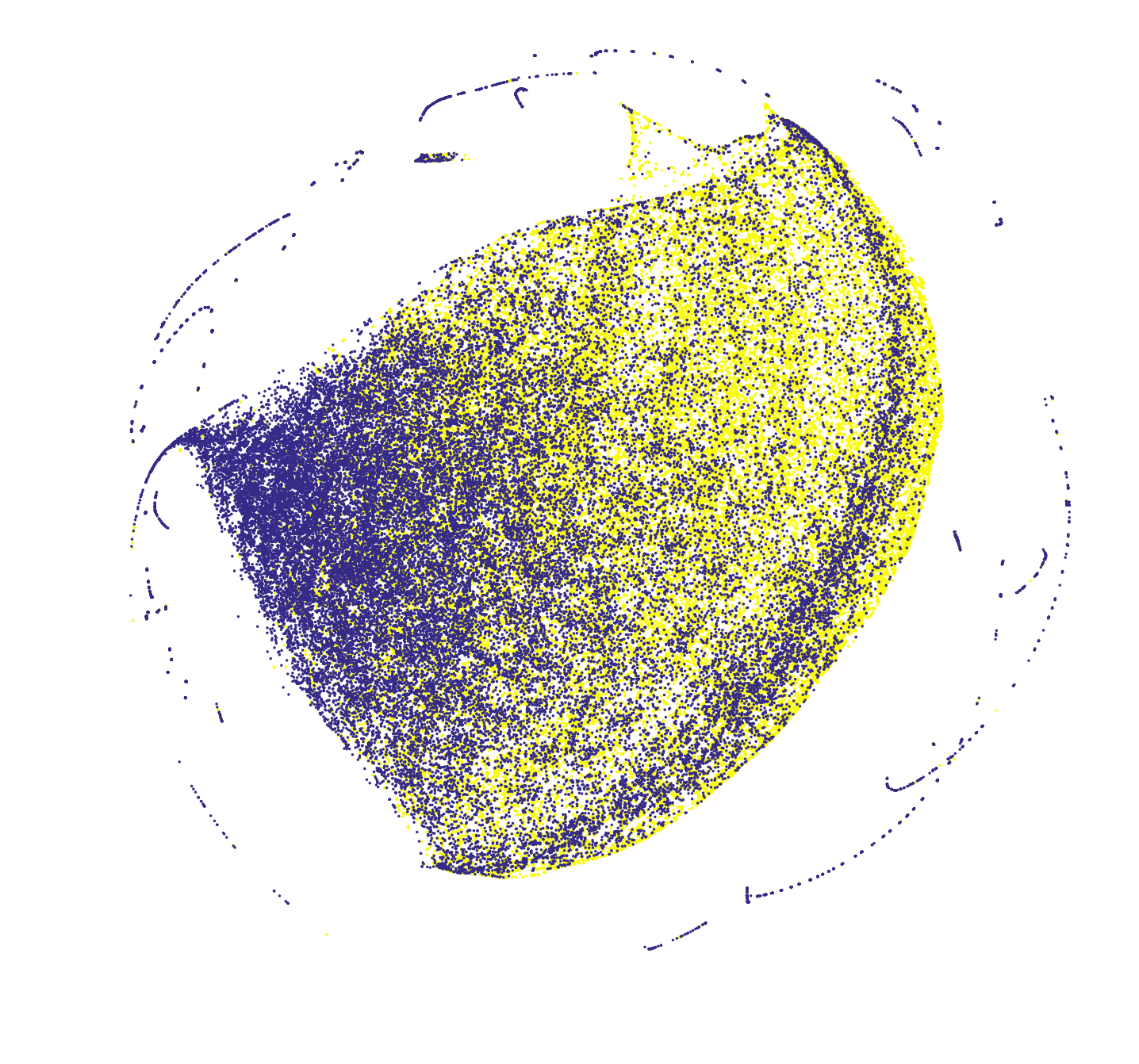}\hfill\\
\end{tabular}
\caption{\label{fig:tvnews} TV News: commercial (yellow), non-commercial (blue) broadcasts. TriMap finds a smooth map while the other two methods split the data into smaller patches.}
\end{center}
\vspace{-0.6cm}
\end{figure*}

\subsection{Effect of changing $t$ and $t'$}
\label{sub:t}
We show the effect of changing the parameters $t$ and $t^\prime$ in 
Figure~\ref{fig:mit-t1}. The experiments are performed on the USPS Digits dataset~\footnote{\url{https://cs.nyu.edu/~roweis/data.html}}. In the first set of experiments, we fix $t=1$ (i.e. use $\log$-transformation) and increase the value of $t^\prime = 2$ (i.e. the tail-heaviness of the similarity function). It is clear from the results that having heavier-tail provides more separability but cannot fix the clutter introduced by the unsatisfied triplets. Next, we fix the similarity function to Student t-distribution ($t^\prime = 2$) and increase the value of $t$ for the $\log_t$-transformation. Value of $t=0$ corresponds to no transformation (linear loss) and $t=1$ recovers the $\log$-transformation. Note that larger values of $t$ provides denser clusters and more separate clusters and reduces the clutter introduced by unsatisfied triplets. We use $t=t^\prime = 2$ in our experiments.

\section{Experiments}

\begin{figure*}[th!]
\begin{center}
\begin{tabular}{m{0.25\textwidth} m{0.25\textwidth} m{0.25\textwidth}}
\multicolumn{1}{c}{t-SNE} & \multicolumn{1}{c}{LargeVis} &  \multicolumn{1}{c}{TriMap}\\
\includegraphics[width=0.25\textwidth]{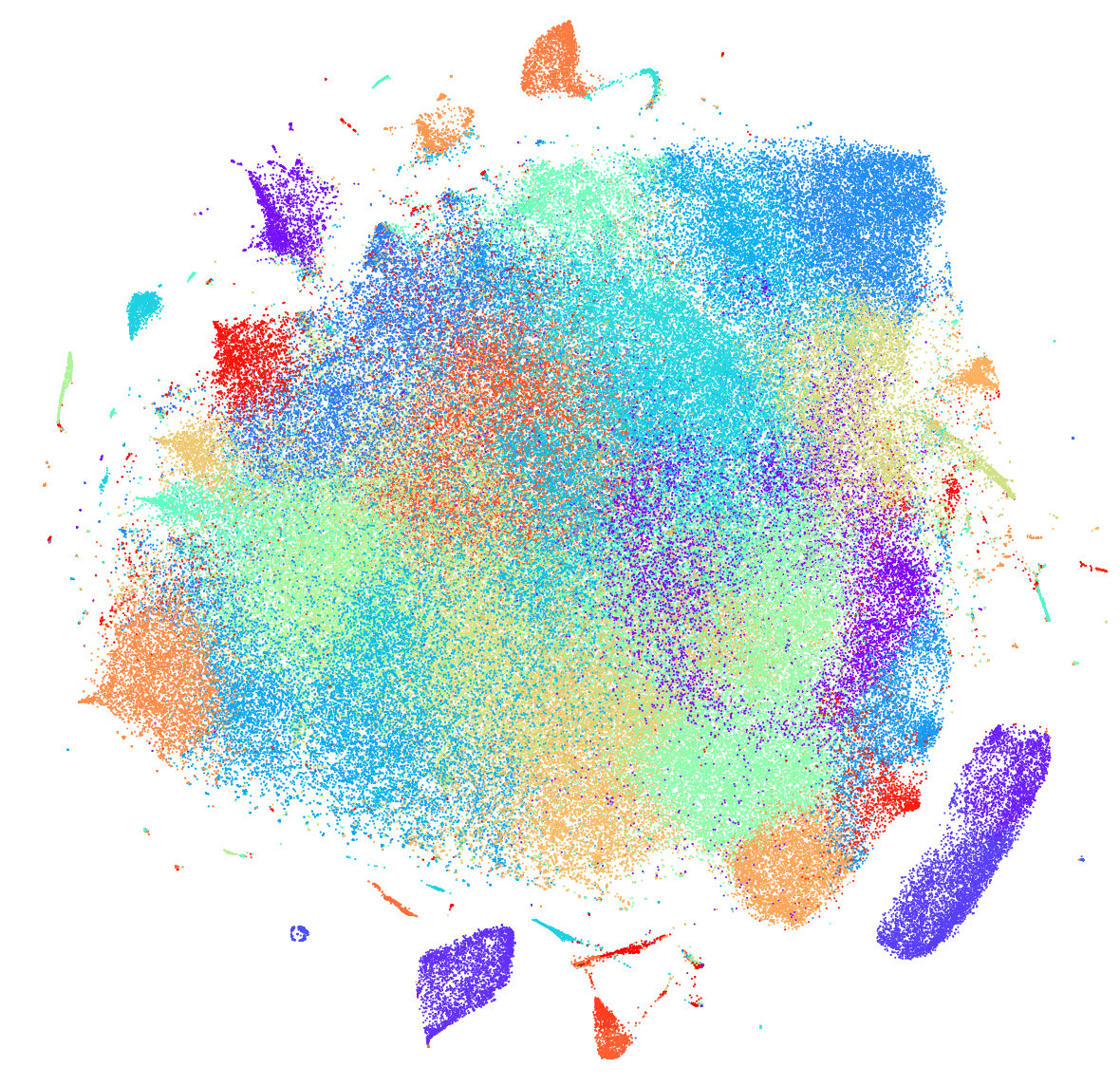} & \includegraphics[width=0.25\textwidth]{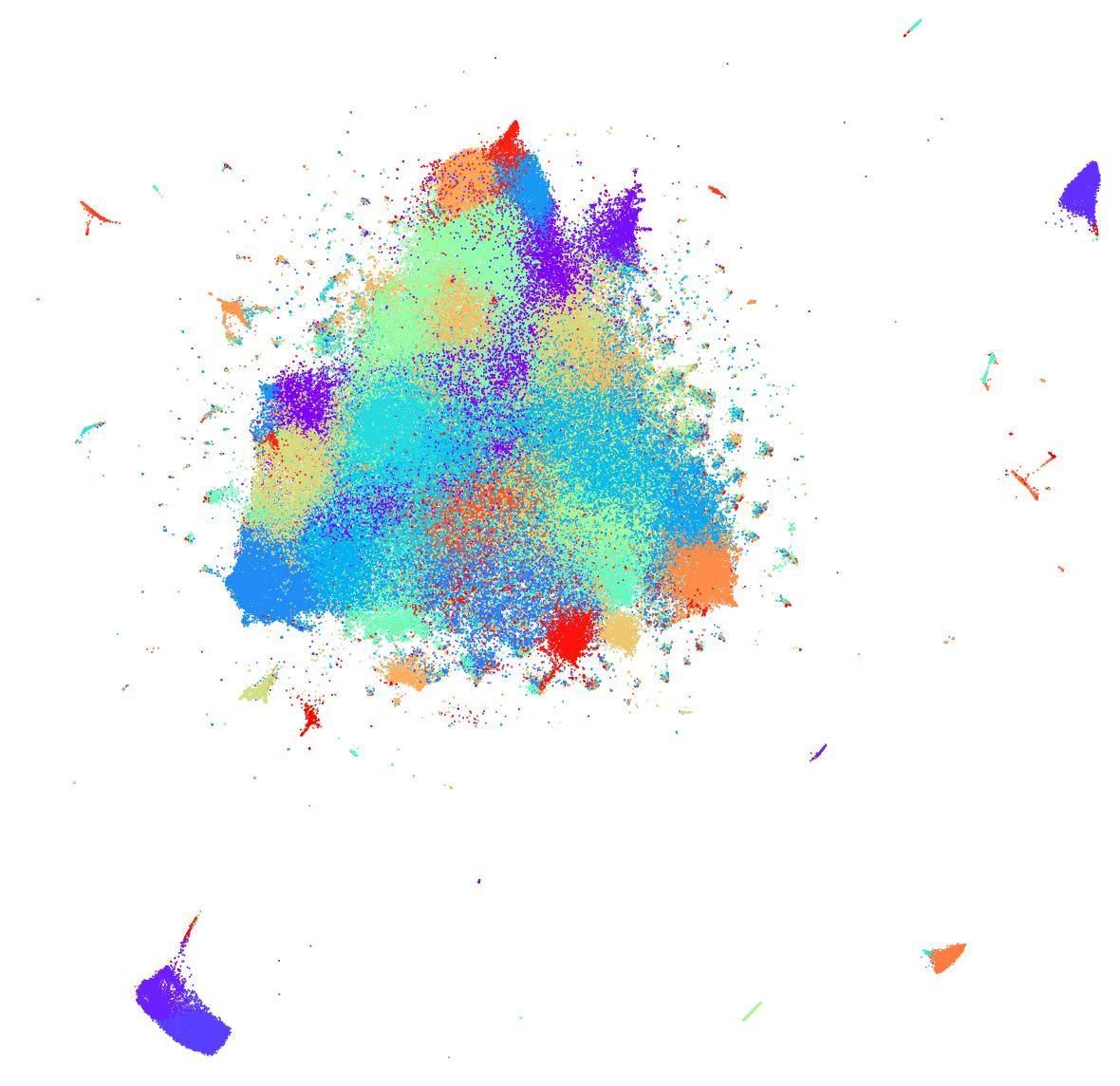} &   \includegraphics[width=0.25\textwidth]{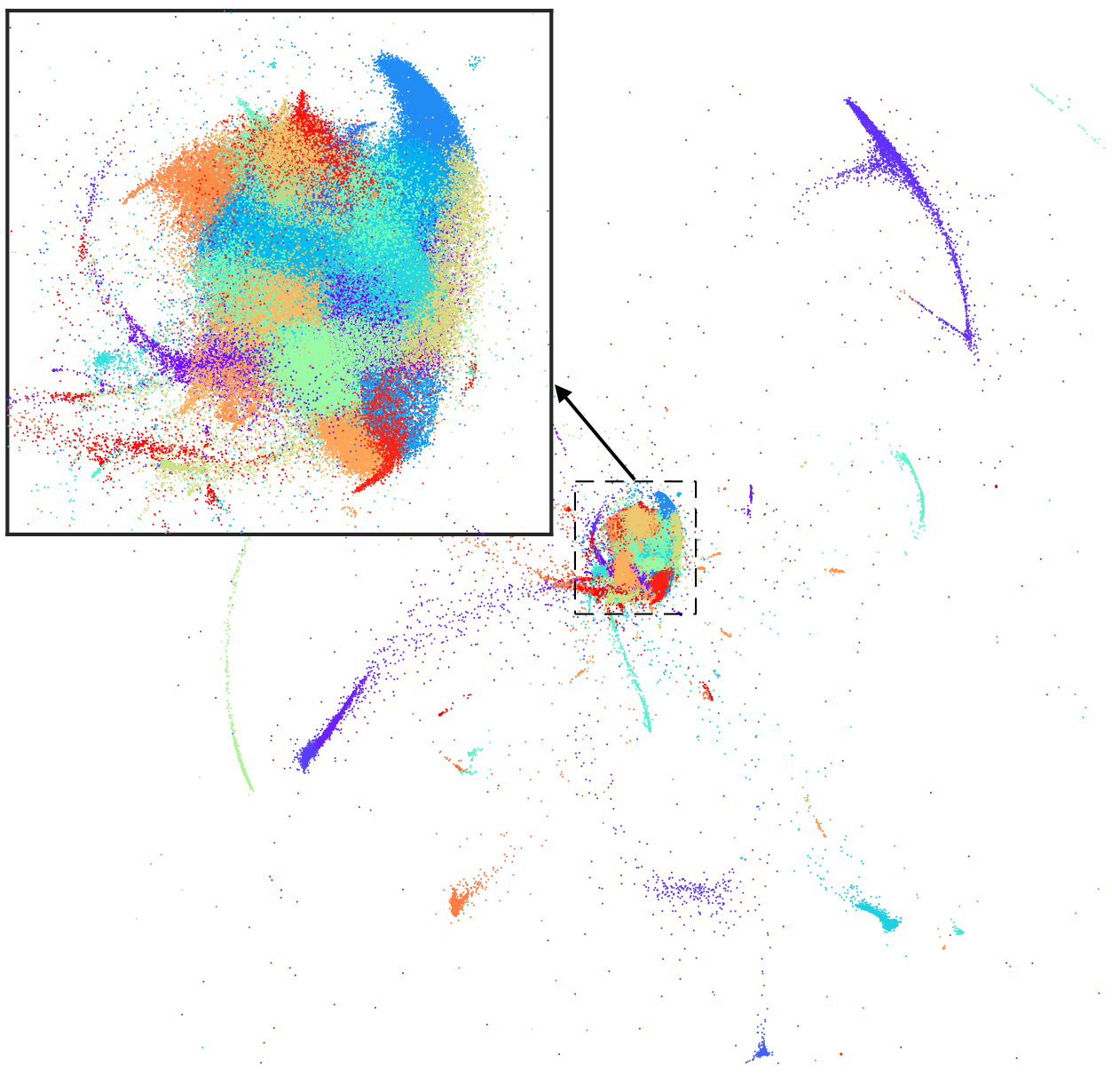}\hfill\\
\end{tabular}
\caption{\label{fig:lyrics} 380k+ lyrics: TriMap recovers multiple scales and outliers. The larger cluster found by both t-SNE and LargeVis is shown in the middle.}
\end{center}
\vspace{-0.5cm}
\end{figure*}

\begin{figure*}[h!]
\vspace{-2mm}
\begin{center}
\begin{tabular}{m{0.25\textwidth} m{0.25\textwidth} m{0.25\textwidth}}
\multicolumn{1}{c}{t-SNE} & \multicolumn{1}{c}{LargeVis} &  \multicolumn{1}{c}{TriMap}\\
\includegraphics[width=0.25\textwidth]{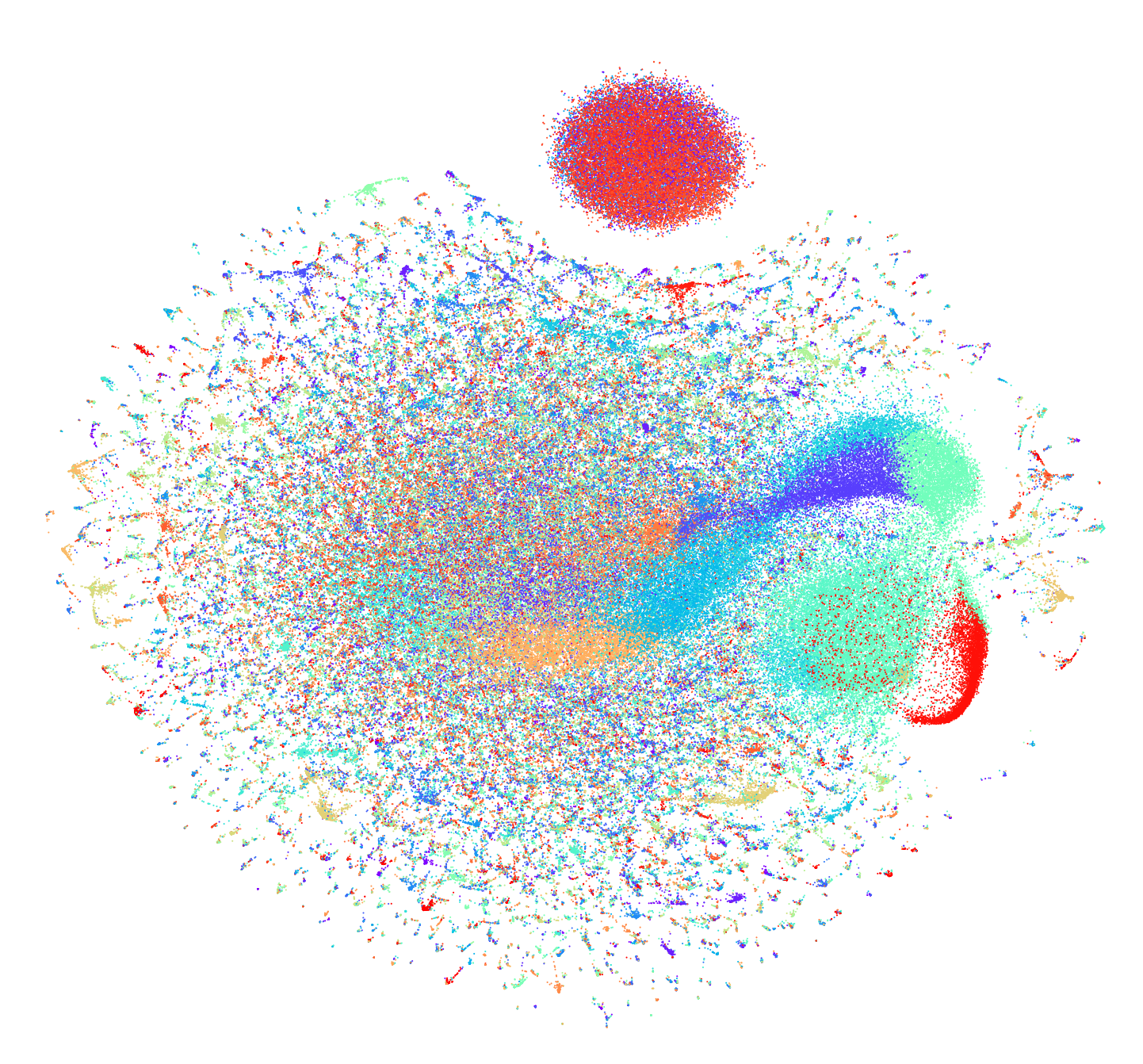} & \includegraphics[width=0.25\textwidth]{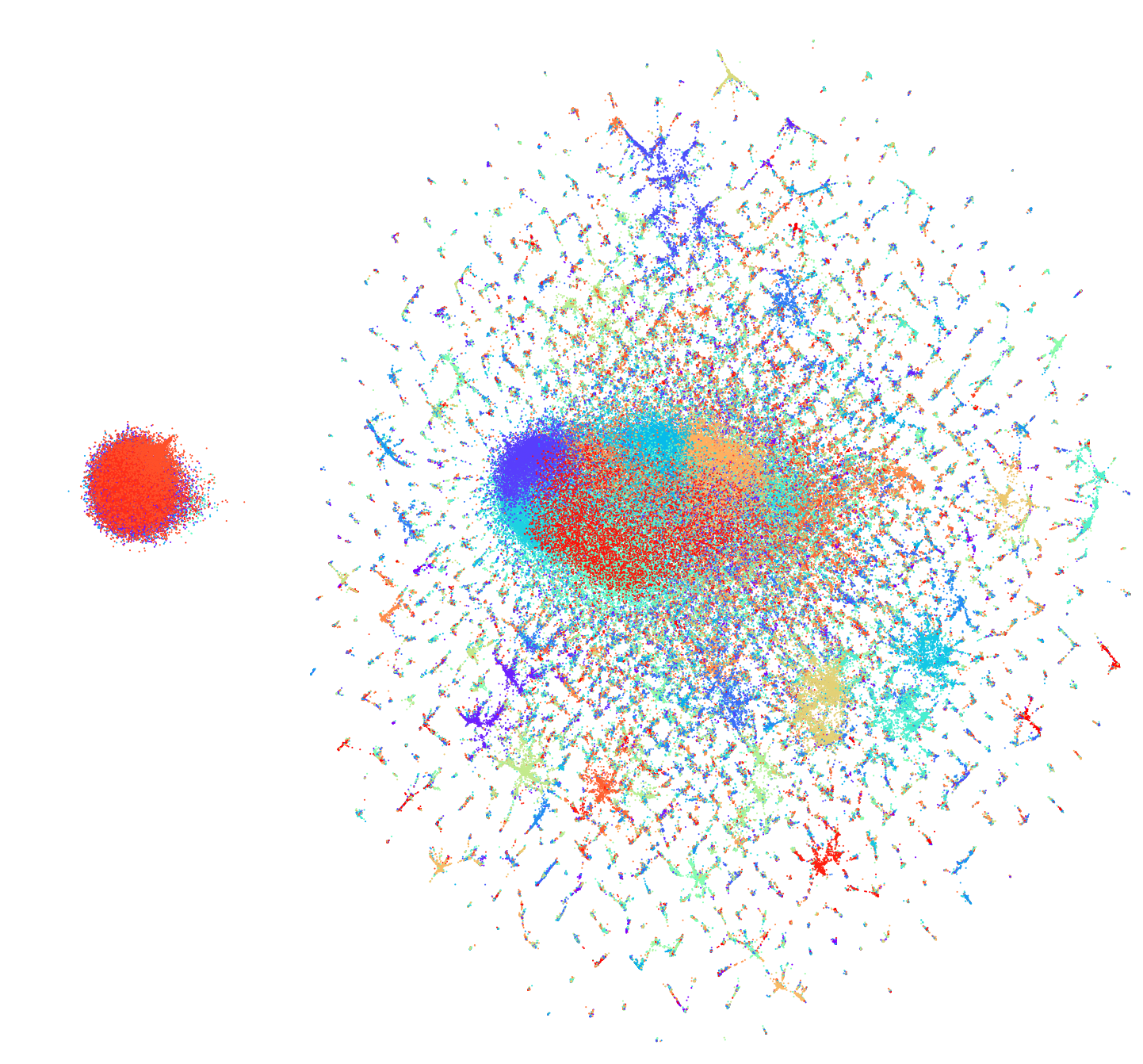} &   \includegraphics[width=0.25\textwidth]{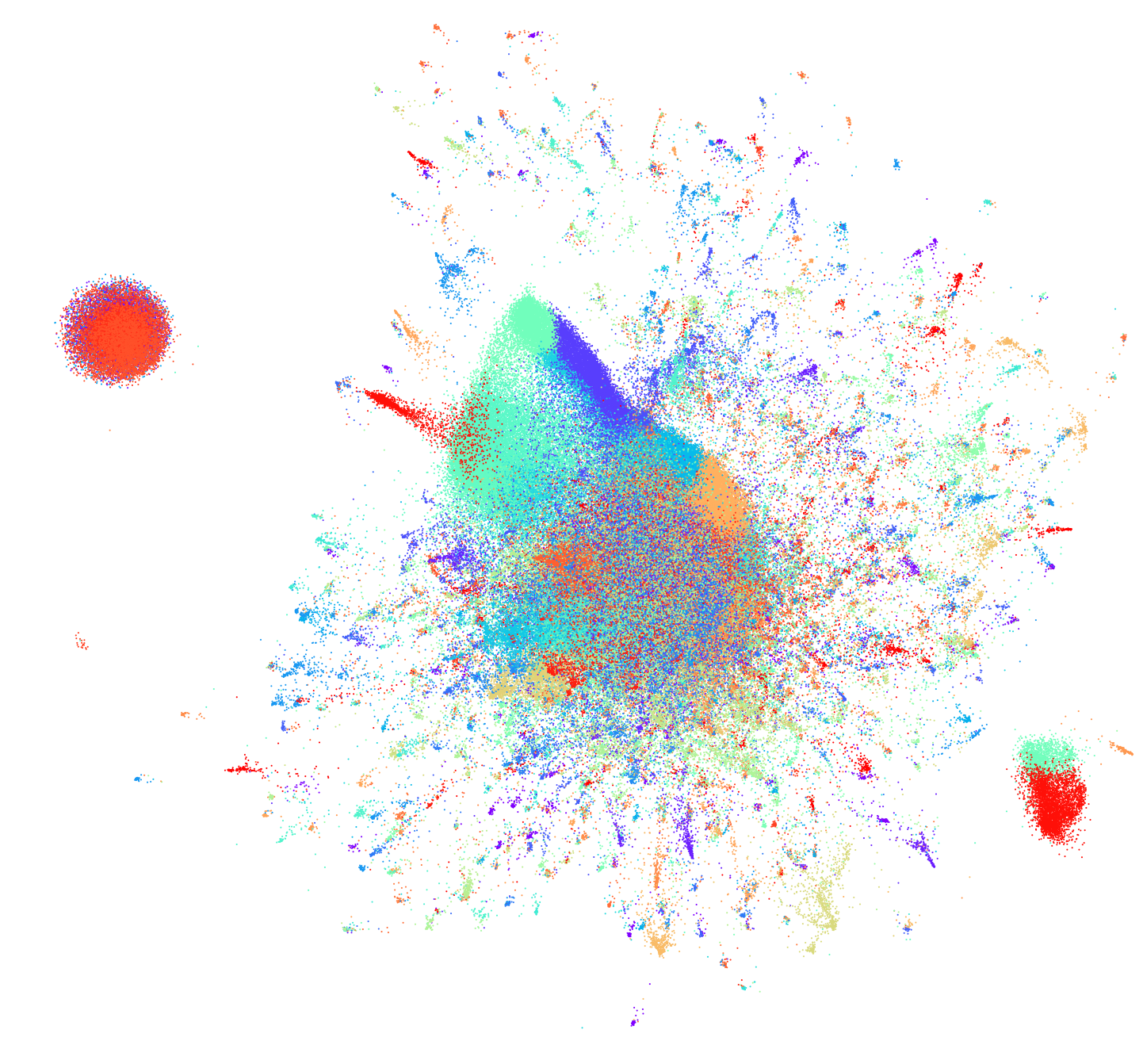}\hfill\\
\end{tabular}
\caption{\label{fig:dblp} DBLP dataset: collaboration network of authors. Both LargeVis and TriMap produce compelling results. However, LargeVis tends to groups the outliers together into small clusters.}
\end{center}
\vspace{-0.6cm}
\end{figure*} 

We evaluate the performance of TriMap for DR on a number of real-world datasets and compare the results to t-SNE and LargeVis. The code for TriMap as well as all the experiments will be available online. In all our experiments, 
we use the default values $t=t^\prime=2$. We provide DR results on five datasets: 1) Epileptic Seizure Recognition\footnote{\url{https://archive.ics.uci.edu/ml/datasets/Epileptic+Seizure+Recognition}}: contains \num{11500} EEG recordings of patients under $5$ different conditions. Each datapoint corresponds to one second of measurements with $178$ samples (dimensions), 2) Fashion MNIST\footnote{\url{https://github.com/zalandoresearch/fashion-mnist}}: consists of \num{70000} examples. Each example is a $28\times 28$ gray scale image, front $10$ classes of clothing items, 3) TV News\footnote{\url{http://archive.ics.uci.edu/ml/datasets/tv+news+channel+commercial+detection+dataset}}: contains audio-visual features extracted from $150$ hours of TV broadcast from $5$ different TV channels. The dataset contains \num{129685} instances, labeled as commercial/non-commercial, and \num{4120} features. We reduce the number of dimensions to $50$ using PCA before applying the methods, 4) 380k+ lyrics\footnote{\url{https://www.kaggle.com/gyani95/380000-lyrics-from-metrolyrics/data}}: the dataset contains lyrics of 380k+ different songs, gathered from \url{http://www.metrolyrics.com}. We first form a weighted co-occurrence graph for the words based on the frequency of the pairs of words which occur together in a sliding window of length 5 words. Next, we compute the representation of each word using the LINE method~\cite{line}. We use number of negative samples equal to 5 and a learning rate of $0.025$ with number of training samples equal to \num{1000}M. We map each word to a $256$ dimensional vector and then, calculate the representation of each song by a weighted average of the words that occur in the song. After removing the stop words and the songs with no available lyrics, we get \num{266557} instances, 5) DBLP collaboration network\footnote{\url{https://snap.stanford.edu/data/com-DBLP.html}}: the collaboration network of \num{317080} authors. Two authors are connected in the network if they publish at least one paper together. We use LINE to find a \num{256} dimensional representation of the network before applying the methods. We set number of negative samples to 5, the learning rate to $0.025$, and the number of training samples to \num{3000}M.

In our experiments, we use fast t-SNE for speed up. We set $\theta = 0.5$ and use the default parameters and learning rates. For LargeVis, we use the default setting of the learning equal to $1.0$ and number of negative samples equal to $5$. For TriMap, we reduce the dimensionality of data to $50$ if necessary, using PCA and use the Balltree algorithm to calculate the $k$-NN. However, any other fast approximate $k$-NN algorithms such as the one proposed in~\cite{largevis} can be used for further speedup. We use $m=50$, $m^\prime = 10$, and $s=5$ for all datasets. The optimization of the algorithm is performed using batch gradient descent with adaptive learning rate. 
We leave the implementation speed up of our method to future work.

Figure~\ref{fig:epileptic} illustrates the results on the Epileptic Seizure Recognition dataset. TriMap shows a smooth transition between measurements under different conditions while both t-SNE and LargeVis fail to preserve the global structure. The results on the Fashion MNIST dataset are shown in Figure~\ref{fig:fmnist}. t-SNE and LargeVis tend to create many spurious clusters. For instance, the cluster of ``Bags'' is divided into multiple smaller sub-clusters. On the other hand, TriMap represents a smooth embedding where each class in concentrated around the same region and there appears to be a smooth transitions among different classes. Figure~\ref{fig:tvnews} shows the results on the TV News dataset. Bothe t-SNE and LargeVis tend to produce several patches of points that do not have a clear structure. TriMap on the other hand shows a smooth transition between the two classes with a few outliers  that are present around the main cluster. Figure~\ref{fig:lyrics} shows the results on the 380k+ lyrics dataset. We use the $50$ clusters in the high-dimensional space found by the k-means++ algorithm as labels. Unlike the other two methods, TriMap reveals multiple layers of structure with several outliers in the dataset. The main cluster that is shown both by t-SNE and LargeVis is also shown in the center of the map with TriMap. Finally, Figure~\ref{fig:dblp} shows the results on the DBLP dataset. As labels, we use the $50$ clusters in the high-dimensional space found by the k-means++ algorithm. The results of LargeVis and TriMap are comparable. However, LargeVis tends to create more small clusters by grouping the outlier points that are close together.

\section{Conclusion}

Dimensionality reduction will remain a core part of Machine
Learning because humans need to visualize their data in
2 or 3 dimensions. We formulated a number of transformations
and outlier tests that any practitioner can use to test
the visualization method at hand. The main existing methods
dramatically fail our tests. We also present a new
DR method that performs significantly better.
This new method called TriMap is based on triplet embedding but
otherwise incorporates a number of redeeming features employed
by the predecessors such as the use of the heavy tailed distributions and sub sampling for obtaining linear time
complexity. Our results on wide range of datasets are quite compelling and suggest that further advances
can be made in this area. Our next goal is to apply TriMap
to data domains where we can test the quality of the obtained visualization
and outlier detection by further physical experiments.

\bibliography{refs}
\bibliographystyle{icml2018}

\end{document}